\documentclass[journal]{IEEEtran}
\usepackage{amsmath,amsfonts}
\usepackage{algorithm}
\usepackage{algpseudocode}
\usepackage{array}
\usepackage[caption=false,font=normalsize,labelfont=sf,textfont=sf]{subfig}
\usepackage{textcomp}
\usepackage{stfloats}
\usepackage{url}
\usepackage{graphicx}
\usepackage{cite} 
\usepackage[T1]{fontenc}
\usepackage{booktabs}
\usepackage[table,xcdraw]{xcolor}
\usepackage{graphicx}
\usepackage{xcolor}
\usepackage{orcidlink}
\usepackage{tikz}
\hypersetup{
colorlinks=true,
linkcolor=[RGB]{36,113,163},
citecolor=[RGB]{36,113,163}
}
\begin{document}

\title{DP-TTA: Test-time Adaptation for Transient Electromagnetic Signal Denoising via Dictionary-driven Prior Regularization}

\author{Meng Yang\hspace{-1.5mm}$^{~\orcidlink{0009-0009-7824-7306}}$, 
Kecheng Chen\hspace{-1.5mm}$^{~\orcidlink{0000-0001-6657-3221}}$,
Wei Luo\hspace{-1.5mm}$^{~\orcidlink{0009-0002-7905-2251}}$, 
Xianjie Chen\hspace{-1.5mm}$^{~\orcidlink{0009-0008-1199-4297}}$, 
Yong Jia\hspace{-1.5mm}$^{~\orcidlink{0000-0003-3165-3349}}$,
Mingyue Wang\hspace{-1.5mm}$^{~\orcidlink{0009-0003-8663-920X}}$, 
Fanqiang Lin\hspace{-1.5mm}$^{~\orcidlink{0000-0003-4884-1944}}$
\thanks{
Manuscript received xxx; revised xxx; accepted xxx. Date of publication xxx; date of current version xxx. Ministry-Province Joint Project under the Ministry of Natural Resources of China (2024ZRBSHZ143), Chengdu “Revelation and Leadership” Science and Technology Project under (2023-JB00-00032-GX). (Corresponding author: Wei Luo, Fanqiang Lin).

M. Yang, F. Lin and Y. Jia are with the College of Mechanical and Electrical Engineering, Chengdu University of Technology, Chengdu 610059, China. K. Chen is with the City University of Hong Kong, Department of Electrical Engineering, Hong Kong, China. W. L is with the Sichuan Provincial Natural Resources Survey and Design Group Co., Ltd., China. X. C is with the Sichuan Provincial Natural Resources Survey and Design Group Co., Ltd, Chengdu University of Technology, China. M. W is with the Massey University, School of Mathematical and Computational Sciences, Auckland, New Zealand. (email: cdut.daoshan@gmail.com, linfq@cdut.edu.cn, jiayong2014@cdut.edu.cn,cs.ckc96@gmail.com, luovitas@qq.com, 979829772@qq.com, wangmmiia9@gmail.com).

Meng Yang and Kecheng Chen contribute equally to this work. }}

\maketitle

\begin{abstract}
Transient Electromagnetic (TEM) method is widely used in various geophysical applications, providing valuable insights into subsurface properties. However, time-domain TEM signals are often submerged in various types of noise. While recent deep learning-based denoising models have shown strong performance, these models are mostly trained on simulated or single real-world scenario data, overlooking the significant differences in noise characteristics from different geographical regions. Intuitively, models trained in one environment often struggle to perform well in new settings due to differences in geological conditions, equipment, and external interference, leading to reduced denoising performance. To this end, we propose the Dictionary-driven Prior Regularization Test-time Adaptation (DP-TTA). Our key insight is that TEM signals possess intrinsic physical characteristics, such as exponential decay and smoothness, which remain consistent across different regions regardless of external conditions. These intrinsic characteristics serve as ideal prior knowledge for guiding the TTA strategy, which helps the pre-trained model dynamically adjust parameters by utilizing self-supervised losses, improving denoising performance in new scenarios. To implement this, we customized a network, named DTEMDNet. Specifically, we first use dictionary learning to encode these intrinsic characteristics as a dictionary-driven prior, which is integrated into the model during training. At the testing stage, this prior guides the model to adapt dynamically to new environments by minimizing self-supervised losses derived from the dictionary-driven consistency and the signal one-order variation. Extensive experimental results demonstrate that the proposed method achieves much better performance than existing TEM denoising methods and TTA methods. The source code is publicly available at \url{https://github.com/blackyang-1/DP-TTA}.
\end{abstract}


\section{Introduction}
\IEEEPARstart{T}{ransient} electromagnetic (TEM) method is widely used in various geophysical applications (\textit{e.g.,} geological exploration, groundwater investigation,  environmental monitoring)\textcolor[RGB]{36,113,163}{\cite{9258400}},\textcolor[RGB]{36,113,163}{\cite{7564453}},\textcolor[RGB]{36,113,163}{\cite{8662775}}. Specifically, TEM analysis allows us to study the subsurface characteristics by utilizing the received secondary TEM signal\textcolor[RGB]{36,113,163}{\cite{9511824}},\textcolor[RGB]{36,113,163}{\cite{9713881}}. These signals provide critical insights into subsurface features, making TEM an essential tool for various geophysical applications. Nevertheless, the collected TEM signals are often contaminated by various types of noise, such as environmental noise, system oscillations, electromagnetic interference, impulse noise\textcolor[RGB]{36,113,163}{\cite{QI2022111420}}. These noises significantly degrade the quality of the collected TEM signals, thereby reducing the accuracy of subsurface detection\textcolor[RGB]{36,113,163}{\cite{Ji01062018}},\textcolor[RGB]{36,113,163}{\cite{rs16050806}},\textcolor[RGB]{36,113,163}{\cite{min13081084}}. As a result, TEM signal denoising is crucial for improving the signal quality and ensuring reliable and effective analysis of the subsurface features.

\begin{figure}[!t]
\centering
\includegraphics[width=3.5in]{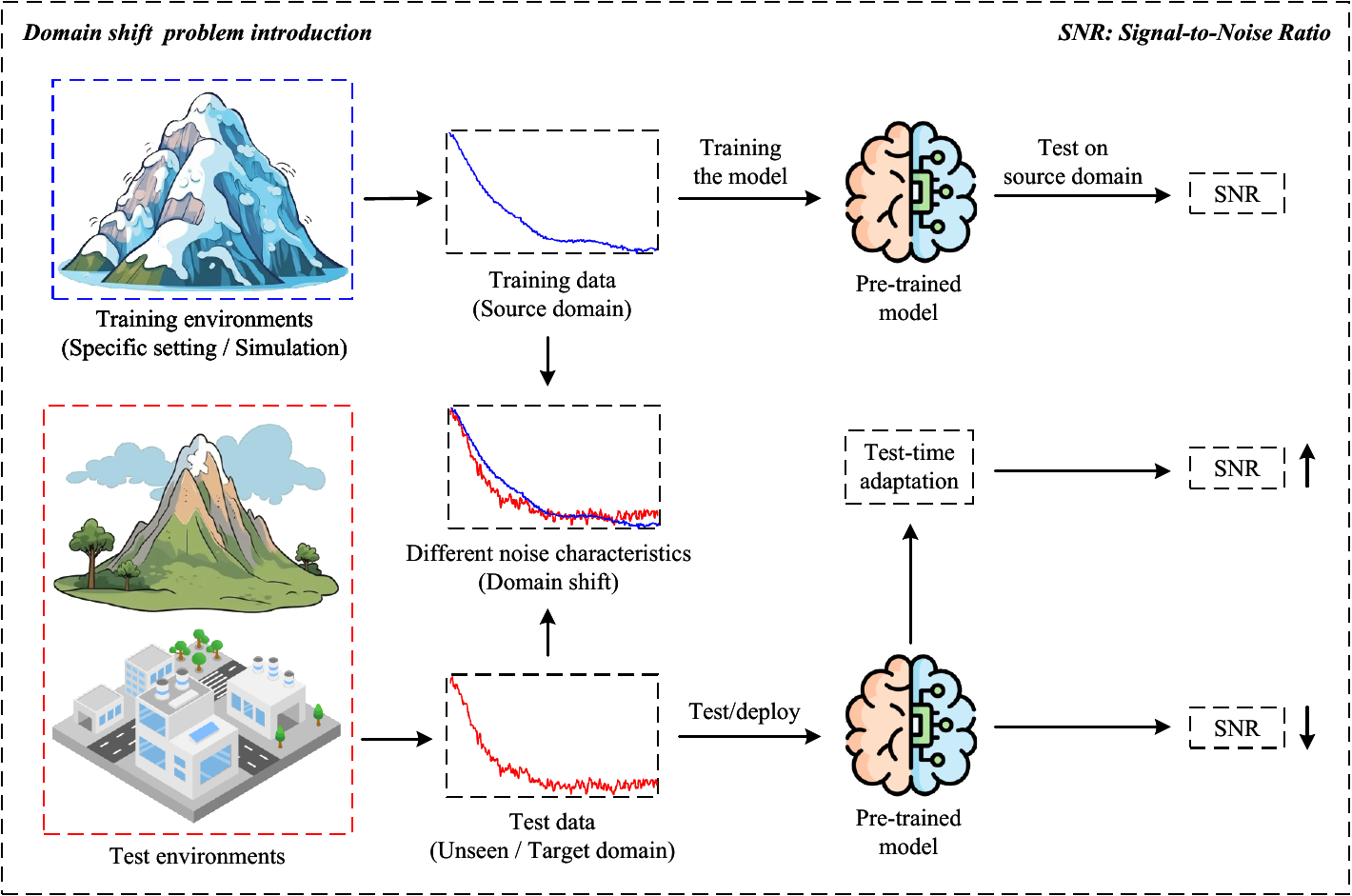}
\caption{Illustration of the domain shift problem. In TEM applications, a domain represents a specific data acquisition environment. For example, high-altitude plateaus, hilly mineral exploration zones, and urban survey areas often exhibit distinct noise characteristics due to differences in geological conditions, equipment, and external interference, etc. Existing models trained on a specific simulation environment (a.k.a., source domain) are difficult to generalize to other environments (a.k.a., target domain), as the discrepancy (i.e., domain shift) of noise characteristics between two domains causes performance degradation. As illustrated in the bottom of the figure, test-time adaptation strategy aims to mitigate this issue by dynamically adjusting the pre-trained model during the testing phase, resulting in better adaptation to new environment with improved denoising performance.}
\label{fig_12}
\end{figure}

Recently, Deep Neural Network (DNN) methods have shown superior performance and have gained widespread attention due to their high denoising capacity and efficient data processing\textcolor[RGB]{36,113,163}{\cite{CHENG2025116494}},\textcolor[RGB]{36,113,163}{\cite{9698089}}. In general, current advances in DNN-based TEM denoising field mainly focus on enhancing denoising capacities using sophisticated model structures, such as SFSDSA\textcolor[RGB]{36,113,163}{\cite{lin2019denoising}} and TEM1Dformer\textcolor[RGB]{36,113,163}{\cite{pan2023tem1dformer}}. Although these approaches exhibit impressive performance on specific simulated and real-world measurement areas, we argue that these models lack sufficient denoising abilities across different environments as they implicitly follow an independent identically distributed (i.i.d.) assumption over the noise distribution between the training and testing data. In the real-world, such an ideal i.i.d. assumption usually violates. For example, as illustrated in Fig. \textcolor[RGB]{36,113,163}{\ref{fig_12}}, the noise characteristics of TEM signals vary significantly across different environments due to differences in geological conditions, equipment, and external interference, etc. When existing denoising models are trained on a specific environment, referred to as the source domain (\textit{e.g.,} a real-world field setting or a simulation dataset), these pre-trained models suffer from degraded denoising performance on new environments (referred to as the target domain or unseen domain), wherein the noise distribution of the source domain will be different from that of the target domain\textcolor[RGB]{36,113,163}{\cite{liang2025comprehensive}}. Such a so-called domain shift issue is quite urgent and troublesome for actual deployments of modern DNN-based TEM denoising models\textcolor[RGB]{36,113,163}{\cite{liang2025comprehensive}},\textcolor[RGB]{36,113,163}{\cite{Zhang_2023_ICCV}}. 

To address this domain shift problem, one of the feasible approaches is to conduct test-time adaptation (TTA), which enables the model (pretrained on a specific environment or simulated dataset) to adapt to test data at the inference (test) time without the requirement of supervised information on the target domain\textcolor[RGB]{36,113,163}{\cite{chen2025gradient}},\textcolor[RGB]{36,113,163}{\cite{dong2024medical}},\textcolor[RGB]{36,113,163}{\cite{zhang2024pass}}. While recent TTA strategies have shown promising results across tasks such as classification\textcolor[RGB]{36,113,163}{\cite{xiao2024modeladaptationtesttime}} and segmentation\textcolor[RGB]{36,113,163}{\cite{chen2024each}},\textcolor[RGB]{36,113,163}{\cite{10457051}},\textcolor[RGB]{36,113,163}{\cite{chen2024DGSM}}, we argue that scaling them to TEM signal denoising tasks may not be as effective\textcolor[RGB]{36,113,163}{\cite{adachi2024test}}. First, many TTA designs are tailored for high-level semantic tasks (\textit{e.g,} classification, object detection, and segmentation) and may not be directly transferable to TEM signal denoising\textcolor[RGB]{36,113,163}{\cite{10843248}},\textcolor[RGB]{36,113,163}{\cite{Ruan_2024_CVPR}}, where the input data exhibit more predictable structures, making them less adaptable to the specific challenges of signal denoising. Second, most of these methods do not fully consider the unique characteristics of TEM signals, such as exponential decay. The absence of mechanisms to leverage these specific signal characteristics may lead to suboptimal performance, where precise handling of such features is crucial for effective denoising.

To this end, we propose Dictionary-driven Prior Regularization Test-time Adaptation (DP-TTA).  In our method, the core motivation stems from the observation that TEM signals exhibit domain-invariant intrinsic characteristics across measurement areas or environments, such as exponential decay and smoothness signal properties, which can serve as ideal prior knowledge to guide the pre-trained model for dynamic parameter optimization by minimizing these prior-driven self-supervised losses. Specifically, by introducing the noisy target-domain TEM signal and its data-augmentation signal at test time, DP-TTA framework enforces the denoising model to be consistent over these two views in terms of exponential decay and smoothness signal properties, which can be reflected by a dictionary-driven consistency loss and a signal one-order variation-driven self-supervised loss, respectively. \textit{Although such consistency regularization has shown its effectiveness in previous TTA tasks}\textcolor[RGB]{36,113,163}{\cite{ruggeri2021treeconstrainedgraphneuralnetworks}} \textit{due to its robustness encouragement for the model, this study is the first to consider domain-invariant intrinsic characteristics of TEM signals as effective consistency cues.} To implement the dictionary-driven consistency loss, we first leverage the dictionary learning to capture the exponential decay property of TEM signals on the training dataset (source domain) and then enforce the model to reconstruct consistent outputs based on the learned dictionary at the test time. Meanwhile, we use the one-order variation between signal points to characterize the smoothness of denoised TEM signals, leading to an effective self-supervised loss at the test time.
The contributions of this paper are fourfold:
\begin{itemize}
    \item This paper is the first to address the domain shift problem of TEM signal denoising from a test-time perspective. To this end, a novel TTA framework, namely DP-TTA, is proposed to enable an effective adaptation of the pretrained denoising model on new environments (i.e., target domains), leading to better denoising performance.

    \item A novel dictionary-driven consistency loss is proposed to introduce the intrinsic exponential decay property of TEM signals as a consistency cue, resulting in better robustness to domain shift.
    
    \item A novel signal one-order variation-driven self-supervised loss is proposed to encourage the model to preserve the smoothness of TEM signals, strengthening the ability to suppress subtle perturbations and achieving better test-time adaptation performance.

    \item DP-TTA is validated on different domain shift problems. Extensive experiments demonstrate that our method outperforms both existing TEM signal denoising methods and TTA strategies, showcasing the potential of our approach in real-world TEM signal denoising tasks.
\end{itemize}

\section{Related Work}
In this section, we review and discuss TEM signal denoising methods and the existing Test time adaptation methods.
\subsection{Traditional TEM Signal Denoising Methods}
Traditional methods, such as Kalman filtering\textcolor[RGB]{36,113,163}{\cite{grewal2014kalman}}, wavelet transforms\textcolor[RGB]{36,113,163}{\cite{hu2021signal}}, empirical mode decomposition (EMD)\textcolor[RGB]{36,113,163}{\cite{rilling2003empirical}}, variational mode decomposition (VMD)\textcolor[RGB]{36,113,163}{\cite{nazari2020successive}}, have been proven to effectively remove noise while preserving the structure of the signal\textcolor[RGB]{36,113,163}{\cite{Ji01062018}},\textcolor[RGB]{36,113,163}{\cite{9600685}}. Currently, most traditional methods adopt automatic parameter optimization algorithms to avoid the need for manual tuning. For example, Wei et al.\textcolor[RGB]{36,113,163}{\cite{9810825}} proposed the SMA-VMD-WTD denoising algorithm, which uses the slime mould algorithm (SMA) to automatically optimize the subsequent denoising process, achieving an autonomous denoising process. Similarly, Feng et al.\textcolor[RGB]{36,113,163}{\cite{FENG2021109815}} used the whale optimization algorithm (WOA) to quickly obtain optimal VMD parameters for signal decomposition, then employed the Bhattacharyya distance algorithm to identify clean signal and noise modes, achieving noise suppression and TEM signal reconstruction. Moreover, Yan et al.\textcolor[RGB]{36,113,163}{\cite{10322753}} and Li et al.\textcolor[RGB]{36,113,163}{\cite{10843714}} also leveraged similar parameter optimization ideas. Overall, traditional algorithms have made progress by eliminating the need for manual parameter tuning, but when faced with big data, these methods often suffer from high time and computational resource consumption\textcolor[RGB]{36,113,163}{\cite{101093}}.

\subsection{DNN-based TEM Signal Denoising Methods}
In recent years, DNN-based methods have been widely applied in fields such as computer vision and signal processing\textcolor[RGB]{36,113,163}{\cite{He_2016_CVPR}},\textcolor[RGB]{36,113,163}{\cite{He_2022_CVPR}},\textcolor[RGB]{36,113,163}{\cite{sun2025noiseconditioningnecessarydenoising}},\textcolor[RGB]{36,113,163}{\cite{chen2024deconstructingdenoisingdiffusionmodels}}. Benefiting from this, DNN-based TEM signal denoising methods have also seen significant development. For example, Chen et al.\textcolor[RGB]{36,113,163}{\cite{9258400}} proposed TEMDNet, an end-to-end denoiser that handles arbitrary scales and demonstrates excellent and fast denoising performance. Wang et al.\textcolor[RGB]{36,113,163}{\cite{9698089}} introduced TEM-NLNet, which utilizes generative adversarial networks (GANs) for better denoising of real TEM signals, though it incurs high training time costs. Pan et al.\textcolor[RGB]{36,113,163}{\cite{pan2023tem1dformer}} combined 1D convolutions with Vision Transformers (ViT) to address overfitting in 1D sequence denoising tasks. Nevertheless, while existing DNN-based methods demonstrate superior performance, they generally lack abilities to handle domain shift problem, which severely limits their generalizability in real-world deployments. As a result, a more efficient and simplified approach is needed to tackle this issue.

\subsection{Test-time Adaptation (TTA)}
TTA has emerged as a promising solution to address domain shift by enabling pre-trained models to adapt to unseen domains without requiring access to labeled data. Currently, prevailing TTA strategies are mostly designed to solve high-level tasks, leveraging surrogate objectives such as entropy minimization or feature alignment to facilitate adaptation to the target domain\textcolor[RGB]{36,113,163}{\cite{Wang_2023_CVPR}},\textcolor[RGB]{36,113,163}{\cite{wang2020tent}}. For instance, Chen et al.\textcolor[RGB]{36,113,163}{\cite{Chen_2022_CVPR}} proposed contrastive test-time adaptation, which adjust the model parameters through the provision of low-noise pseudo-labels, leading to more reliable adaptation to the target domain. Fahim et al.\textcolor[RGB]{36,113,163}{\cite{fahim2023ss}} proposed a versatile TTA-based image denoising framework that enforces prediction consistency based on real-world constraints, demonstrating improvements in both supervised and unsupervised regimes. Similarly, Mansour et al.\textcolor[RGB]{36,113,163}{\cite{101007}} introduced a TTA strategy that incorporates masked image modeling and corresponding reconstruction losses at both training and inference stages, outperforming zero-shot baselines. While these approaches exhibit methodological novelty, they are fundamentally driven by data-dependent adaptation objectives, lacking the utilization of explicit priors. Besides, they are not specifically tailored for signal denoising problems. As a result, we hypothesize that directly scaling these methods to the TEM signal denoising domain may be suboptimal.

\section{Methodology}
We begin by formulating the theoretical denoising problem of TEM signals and analyzing the domain shift problem from a Bayesian perspective, establishing the theoretical basis for the motivation of DP-TTA. Then, we present the intrinsic characteristics of TEM signal and propose the DP-TTA. Next, we introduce the self-supervised framework and detail the implementation of the dictionary learning process, the model structure, and the hyperparameter settings of DP-TTA.

\subsection{Preliminary and Theoretical Motivation}
Typically, the observed TEM signal \( y(t) \) is modeled as a superposition of the clean signal and additive noise, can be described by the following equation\textcolor[RGB]{36,113,163}{\cite{9258400}},\textcolor[RGB]{36,113,163}{\cite{9698089}},\textcolor[RGB]{36,113,163}{\cite{Huang_2025}}:

\begin{equation}
y(t) = v(t) + \epsilon(t),
\end{equation}
where \( v(t) \) represents theoretical, noise-free TEM response, and \( \epsilon(t) \) accounts for measurement noise. In geophysical contexts, the clean signal \( v(t) \) is commonly modeled as an exponentially decaying function of time, described by the following forward model:

\begin{equation}
s = v(t) = \frac{C}{\tau} \sum_{k=1}^{\infty} \exp\left(-\frac{k^2 t}{\tau}\right) + B,
\end{equation}
where \( C \) is a constant dependent on subsurface characteristics (\textit{e.g,} the depth of the conductor, the radius of the transmitting coil), and \( \tau \) is the time constant of the conductor. The term \( B \) represents a bias associated with the direct current offset. As in\textcolor[RGB]{36,113,163}{\cite{9258400}}, the actual noisy TEM signal can be written as

\begin{equation}
y(t) = s + \epsilon(t).
\end{equation}

We propose to tackle the TEM denoising problem on the basis of the aforementioned forward degradation model, as this task itself constitutes an ill-posed inverse problem. In real-world scenarios, the noise affecting TEM signals is often a combination of multiple noise sources, each with distinct characteristics\textcolor[RGB]{36,113,163}{\cite{1093}},\textcolor[RGB]{36,113,163}{\cite{9810825}}. The noise \( \epsilon(t) \) in actual TEM signal acquisition can include:

\begin{itemize}
    \item Random Environmental Noise: This noise is typically modeled as Gaussian noise \( \epsilon(t) \sim \mathcal{N}(0, \sigma^2) \), arising from uncontrollable environmental fluctuations.
    \item System Oscillations: Low-frequency disturbances induced by mechanical vibrations or internal system instabilities.
    \item Electromagnetic Interference (EMI): A common source of high-frequency noise in TEM signals, originating from nearby electronic devices, transformers, or power supplies. 
    \item Impulse Noise: Noise caused by sudden, brief, high-amplitude disturbances, such as power surges or malfunctioning electronic components.
\end{itemize}

These diverse and heterogeneous noise sources substantially increase the difficulty of the denoising task. More importantly, their variability and unpredictability amplify the risk of distributional mismatch between the source domain and the target domain, exacerbating the domain shift problem.

Here, we provide a Bayesian perspective to explain the domain shift problem\textcolor[RGB]{36,113,163}{\cite{chapeau2004noise}}\textcolor[RGB]{36,113,163}{\cite{10167840}},\textcolor[RGB]{36,113,163}{\cite{4360034}}. In the Bayesian framework, TEM signal denoising can be understood as maximizing the posterior probability \( p(s|y) \)\textcolor[RGB]{36,113,163}{\cite{9258400}}. The posterior probability can be expressed as

\begin{equation}
    p(s|y) \propto p(y|s) p(s),
\end{equation}
where \( p(y|s) \) is the likelihood function, which models the probability of observing the noisy signal \( y \) given the clean signal \( s \), and \( p(s) \) is the prior distribution, which represents the prior knowledge about the clean signal \( s \). To simplify the computation, the logarithm of the posterior probability is typically taken:

\begin{equation}
    \hat{s} = \arg \max_s \left( \log p(y|s) + \log p(s) \right).
\end{equation}
Then, equation (5) is transformed into a minimization objective by taking the negative log-likelihood.

\begin{equation}
\hat{s} = \arg \min_s \left( \frac{1}{2} \| y - s \|^2 + \lambda R(s) \right).
\end{equation}
The first term \( \frac{1}{2} \| y - s \|^2 \) represents the data fidelity term derived from the Gaussian likelihood assumption, and the second term \( R(s) \) is a regularization term reflecting the prior. The parameter \( \lambda \) balances the two terms. As in \textcolor[RGB]{36,113,163}{\cite{bishop2006pattern}},\textcolor[RGB]{36,113,163}{\cite{blei2017variational}}, deep learning-based models can be viewed as implicitly learning both the likelihood and the prior from training data. 

Based on the aforementioned theory, we denote the source and target TEM signals as
\begin{equation}
y_s = s + \epsilon_s, \quad y_t = s + \epsilon_t,
\end{equation}
where \( y_s \) and \( y_t \) are the noisy observations in the source and target domains, and \( s \) is the underlying clean signal. During training, the model is only exposed to source domain data \( p(s | y_s) \propto p(y_s | s) p(s) \). This leads to a denoising objective composed of two components: a data fidelity term \( \| y - s \|^2 \) corresponding to the likelihood \( p(y|s) \), and a regularization term \( R(s) \) encoding the prior \( p(s) \). However, when the model is applied to target domain inputs \( y_t \), where the noise \( \epsilon_t \) differs from \( \epsilon_s \), the inference becomes suboptimal due to two types of mismatches:

\begin{itemize}
    \item \textbf{Likelihood mismatch:} The fidelity term \( \| y - s \|^2 \) assumes the noise distribution matches that of the training data (i.i.d). In the target domain, noise exhibits different patterns, leading the model to rely on a likelihood that no longer reflects the actual observation process, resulting in incorrect reconstructions.
    
    \item \textbf{Prior mismatch:} The regularization term \( R(s) \), while intended to capture the structure of clean TEM signals, is implicitly co-adapted to the source noise context. That is, the model learns a signal prior that is entangled with the noise statistics seen during training. When noise characteristics change, this prior may impose misleading constraints, harming the model’s ability to recover the true signal.
\end{itemize}

As a result, these mismatches result in biased inference on the target domain, leading to degraded denoising performance. This highlights that domain shift problem in TEM signal denoising is not only a problem of input distributional changes, but also a deeper issue of probabilistic inconsistency in the model's learned assumptions.

\subsection{Integration of the Dictionary Learning}

\begin{figure}[!t]
\centering
\includegraphics[width=3.5in]{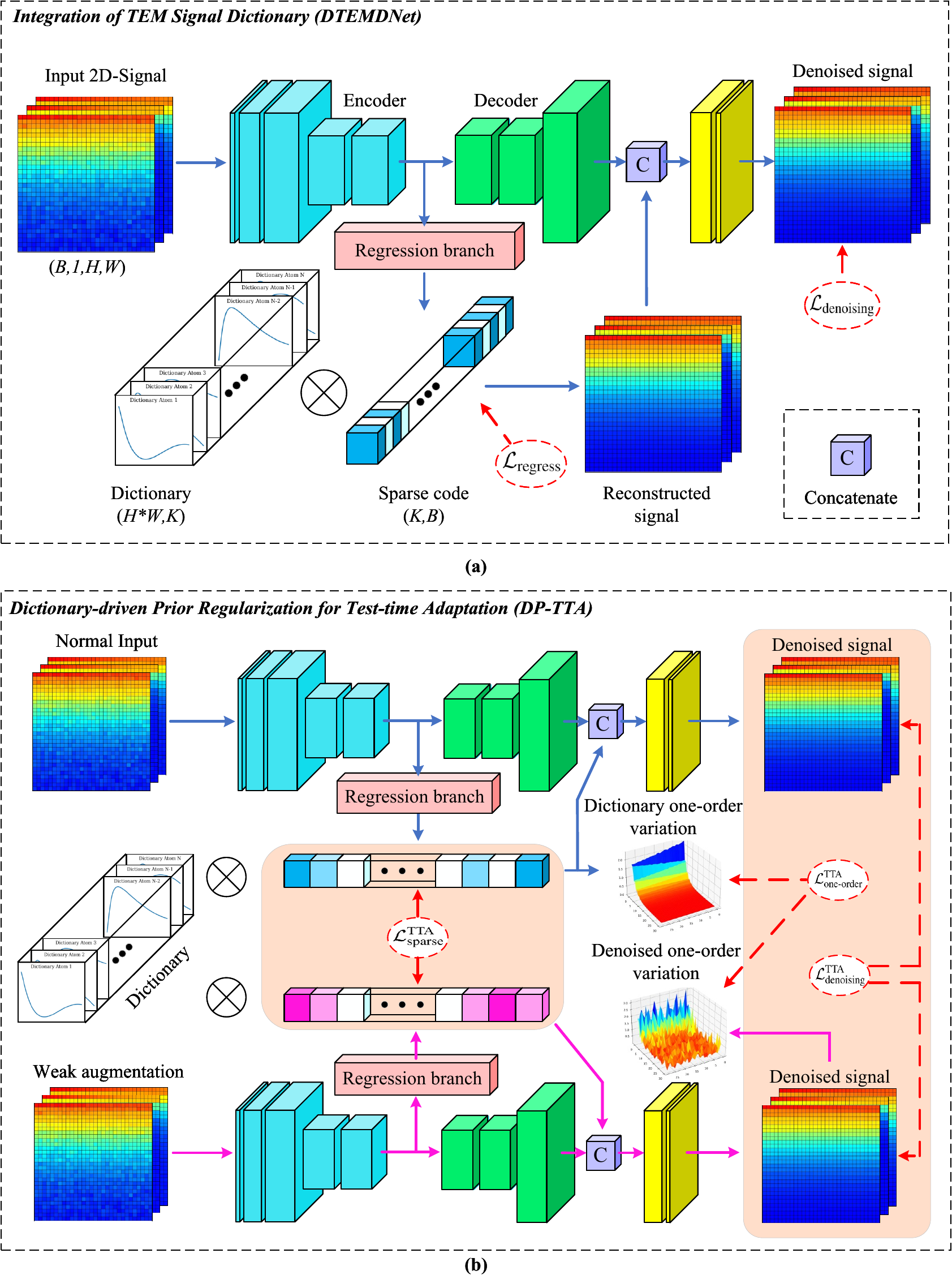}
\caption{Overview of the proposed framework. (a) illustrates the complete DTEMDNet architecture integrating the TEM signal dictionary. In the source-domain training, DTEMDNet is pre-trained using two supervised losses. (b) demonstrates DP-TTA, where the DTEMDNet performs dynamic self-adaptation at test time by applying three self-supervised losses (Denoising result, sparse code, one-order variation). Note: Input 2D-Signal denotes the 2D image obtained using the 1D to 2D conversion method in\textcolor[RGB]{36,113,163}{\cite{9258400}}. The detailed definitions of the above losses are provided in later sections.}
\label{fig_1}
\end{figure}

As illustrated in Fig. \textcolor[RGB]{36,113,163}{\ref{fig_1}(a)}, the customized DTEMDNet integrates two complementary branches: the CNN branch and the dictionary learning branch. The encoder first encodes the input TEM signals into latent representations, which are then simultaneously fed into the regression branch and the decoder. The regression branch predicts sparse codes, which are then multiplied by the pre-extracted dictionary atoms to reconstruct the clean TEM signal. This dictionary reconstruction acts as a structurally informed prior, augmenting the denoising process. The decoder refines the denoised output by concatenating its intermediate feature maps with the dictionary reconstruction, effectively leveraging both learned features and explicit dictionary-driven priors. During training, DTEMDNet is optimized using a combination of two loss functions. Specifically, the denoising loss $\mathcal{L}_{\text{denoising}}$ quantifies the reconstruction error between the denoised output and the ground truth clean signal, typically formulated as the mean squared error (MSE):

\begin{equation}
\mathcal{L}_{\text{denoising}} = || \hat{Y} - Y_{\text{truth}} ||_2^2.
\end{equation}

Additionally, a regression loss $\mathcal{L}_{\text{regress}}$ ensures that the predicted sparse codes are consistent with the ground truth sparse codes derived from the dictionary learning. This loss is calculated as the L1 norm between the predicted sparse codes $\hat{A}$ and the ground truth sparse codes $A_{\text{truth}}$:

\begin{equation}
\mathcal{L}_{\text{regress}} = || \hat{A} - A_{\text{truth}} ||_1.
\end{equation}

The total loss $\mathcal{L}_{\text{total}}$ combines the denoising and sparse code losses as follows:

\begin{equation}
\mathcal{L}_{\text{total}} = \alpha \cdot \mathcal{L}_{\text{regress}} + \beta \cdot \mathcal{L}_{\text{denoising}},
\end{equation}\

where $\alpha$ and $\beta$ are hyperparameters that balance the contributions of the two losses. In our implementation, $\alpha$ is set to 10 and $\beta$ to 1 to compensate for the typically smaller magnitude $\mathcal{L}_{\text{regress}}$, ensuring balanced optimization.

\subsection{DP-TTA Method}
The core of our approach lies in the observation that TEM signals consistently exhibit intrinsic physical characteristics, which can serve as ideal domain-invariant priors for denoising tasks and guide the TTA strategy, despite the diversity of real-world domains. Building on this insight, we propose DP-TTA, a strategy that dynamically adjusts pre-trained model parameters via dictionary-driven priors. Specifically, as depicted in Fig. \textcolor[RGB]{36,113,163}{\ref{fig_1}(b)}, we first transform the input signal using the 1D-to-2D method outlined in\textcolor[RGB]{36,113,163}{\cite{9258400}}, then processes each sample in two different ways: one for the normal input and the other for the augmented version (\textit{e.g.,} Gaussian perturbation). 
These two inputs allow the model to produce three sets of outputs: the denoised signals, the predicted sparse codes, and the dictionary-reconstructed signals, which enables prediction consistency regularization. The rationale behind this operation is that, despite the variations in the augmented noisy signals, the underlying clean signals and the model's understanding of the intrinsic characteristics should remain consistent. Through this mechanism, we derive three self-supervised losses originated by the dictionary-driven prior: denoising result, sparse code, and one-order variation loss. By minimizing these losses, we enable the pre-trained model to autonomously adapt during the testing stage, effectively mitigating the domain shift problem. We summarize the DP-TTA in Algorithm \textcolor[RGB]{36,113,163}{\ref{alg:DP-TTA}}.

\begin{algorithm}[htbp]
\caption{DP-TTA Strategy}
\label{alg:DP-TTA}
\begin{algorithmic}[1]

\Statex \textbf{Input:} Normal sample \( x \), augmented sample \( \tilde{x} \).
\Statex \textbf{Output:} Denoised signals \( y_1, y_2 \), sparse codes \( A_1, A_2 \), dictionary reconstruction signals \( d_1, d_2 \).
\State Compute model outputs:
\[(y_1, A_1, d_1) = f_{\theta}(x);\]
\[(y_2, A_2, d_2) = f_{\theta}(\tilde{x});\]
\State Calculate self-supervised losses:
\[
\mathcal{L}_{\text{denoising}}^{\text{TTA}} = \| y_1 - y_2 \|_2^2;
\]
\[
\mathcal{L}_{\text{sparse}}^{\text{TTA}} = \| A_1 - A_2 \|_1;
\]
\[
\mathcal{L}_{\text{one-order}}^{\text{TTA}} = \| \nabla d_1 - \nabla y_2 \|_2^2;
\]
\State Aggregate total loss:
\[
\mathcal{L}_{\text{TTA}} = \beta_1 ( \mathcal{L}_{\text{sparse}}^{\text{TTA}} + \mathcal{L}_{\text{one-order}}^{\text{TTA}} ) + \beta_2 \mathcal{L}_{\text{denoising}}^{\text{TTA}};
\]
\State Update model parameters \( \theta \) by minimizing \( \mathcal{L}_{\text{TTA}} \).
\end{algorithmic}
\end{algorithm}

\subsection{Signal Dictionary Construction}

Through dictionary learning, we extract a dictionary and the sparse codes of all samples from the source domain dataset. These sparse intrinsic representations capture the shared consistency and intrinsic physical characteristics to TEM signals. Specifically, we extract a dictionary from the source domain dataset \(S = \{ (\text{clean}_i, \text{noisy}_i) \}_{i=1}^{N}\), where \(N\) denotes the number of paired clean and noisy signals in the source domain dataset. The dictionary \(D = \{ d_1, d_2, \dots, d_K \}\) contains \(K\) atoms, where \(K\) is a user-defined parameter. The optimization objective for dictionary learning consists of two terms: minimizing the reconstruction error to ensure faithful signal representation, and enforcing sparsity to promote compact encoding using a minimal subset of dictionary atoms. The optimization problem is formulated as

\begin{equation}
\arg \min_{D, A} \sum_{i=1}^{N} \left\| y_i - \sum_{j=1}^{K} d_j \alpha_{ij} \right\|_2^2 + \lambda \|\alpha_i\|_1,
\end{equation}
where \(D\) is the dictionary, \( \alpha_i \) is the sparse code for the \(i\)-th signal \(y_i\), and \(\lambda\) is a regularization parameter that balances the reconstruction error and sparsity. 

This optimization is performed iteratively through alternating updates of the sparse code \(A\) and the dictionary \(D\). Specifically, with the dictionary fixed, the sparse codes \(A\) are obtained by minimizing the combined reconstruction and sparsity losses, using the Adam optimizer for updates. Subsequently, the dictionary \(D\) is updated by solving a least-squares problem for each atom, refining them iteratively to minimize the reconstruction error. To ensure that the learned dictionary possesses sufficient representational capacity for generalization to unseen domains, we adhere to two fundamental principles in its design:

\begin{enumerate}
    \item The number of dictionary templates \(K\) is kept significantly smaller than the number of training samples \(N\), i.e., \(K \ll N\), which effectively reduces the risk of overfitting to the source-domain data (e.g., \(K = 64\), \(N = 100,000\) in our implementation).
    \item The sparse code \(A\) must exhibit sufficient sparsity, ensuring that each signal is represented using only a minimal subset of dictionary atoms. This sparsity-driven design enhances the dictionary's capacity to distill the most essential characteristics of the signal while minimizing redundancy and improving robustness across diverse domains.
\end{enumerate}

\subsection{Self-Supervised Framework and Loss Functions}

During test-time, we generate three self-supervised losses: the denoised output loss, the sparse code loss, and the one-order variation loss, which can be seen in Fig. \textcolor[RGB]{36,113,163}{\ref{fig_1}(b)}. The denoising and sparse code losses preserve the core signal structure, while the one-order variation loss provides physical constraints.

\subsubsection{\textit{Denoising Loss}} This loss enforces consistency between the denoised outputs of the two inputs, encouraging stable predictions. It is defined as

\begin{equation}
\mathcal{L}_{\text{denoising}}^{\text{TTA}} = \left\| \hat{Y}_{\text{denoised}}^{\text{norm}} - \hat{Y}_{\text{denoised}}^{\text{aug}} \right\|_2^2,
\label{eq:denoising_loss}
\end{equation}
where \( \hat{Y}_{\text{denoised}}^{\text{norm}} \) and \( \hat{Y}_{\text{denoised}}^{\text{aug}} \) are the denoised outputs corresponding to the normal and augmented inputs, respectively.

\subsubsection{\textit{Sparse Code Loss}} This loss is defined as the difference between the sparse matrices generated by the model. By ensuring the consistency of sparse codes, we preserve the core signal structure, preventing the model from overfitting to the specific noise characteristics of a single input. The sparse code loss is given by

\begin{equation}
\mathcal{L}_{\text{sparse}}^{\text{TTA}} = \left\| \hat{A}_{\text{norm}} - \hat{A}_{\text{aug}} \right\|_1,
\label{eq:sparse_code_loss}
\end{equation}
where \( \hat{A}_{\text{orig}} \) and \( \hat{A}_{\text{aug}} \) are the sparse codes for the normal and augmented inputs.

\subsubsection{\textit{One-order variation Loss}} This loss is motivated by our observation that, although the decoder’s output attains higher SNR, its one-order variation exhibits less smoothness compared to the dictionary reconstruction. The dictionary reconstruction, despite some drift, better preserves the intrinsic smooth decay characteristic of clean TEM signals. In subsequent experiments, we will provide a more detailed and visual explanation of this finding. 
we introduce the one-order variation information of the dictionary-based reconstruction as a self-supervised signal to regularize the model. By minimizing the discrepancy between the one-order variations of the denoised output and the dictionary reconstruction, the model is explicitly guided to preserve the intrinsic smoothness and structural continuity inherent to TEM signals. The one-order variation loss is defined as

\begin{equation}
\mathcal{L}_{\text{one-order}}^{\text{TTA}} = \left\| \nabla \hat{Y}_{\text{dict}}^{\text{norm}} - \nabla \hat{Y}_{\text{denoised}}^{\text{aug}} \right\|_2^2,
\label{eq:gradient_loss}
\end{equation}
where \( \nabla \) denotes the one-order variation operator, \( \hat{Y}_{\text{dict}}^{\text{norm}} \) represents the dictionary reconstruction from normal input, and \( \hat{Y}_{\text{denoised}}^{\text{aug}} \) is the denoised output corresponding to the augmented input.

Finally, we combine these three self-supervised losses into a joint objective, where we assign appropriate coefficients to each term in order to adjust their contributions and find the optimal balance. The total loss \( \mathcal{L}_{\text{TTA}} \) is a weighted combination of these three components:

\begin{equation}
\mathcal{L}_{\text{TTA}} = \beta_1 \cdot ( \mathcal{L}_{\text{sparse}}^{\text{TTA}} + \mathcal{L}_{\text{one-order}}^{\text{TTA}} ) + \beta_2 \cdot \mathcal{L}_{\text{denoising}}^{\text{TTA}},
\label{eq:total_loss}
\end{equation}
where \( \beta_1 \) and \( \beta_2 \) are hyperparameters that control the relative importance of each element.

\subsection{Implementation Details}

In this section, we provide more implementation details of the model structure and the hyperparameter settings of the DP-TTA.

\subsubsection{Model Structure}
DTEMDNet is primarily constructed using residual connections and dilated convolutions as fundamental building blocks, enabling effective feature reuse and receptive field expansion. A detailed architecture is illustrated in Table~\textcolor[RGB]{36,113,163}{\ref{table:DTEMDNet_Architecture}}.

\setlength{\tabcolsep}{15pt}
\begin{table}[htbp]
\centering
\caption{Overall Architecture of DTEMDNet. Note that Dilated-conv refers to the dilated convolution, ResBlockv1 consists of \texttt{1x1} and \texttt{3x3} convolutions, ResBlockv2 consists of two \texttt{3x3} convolution layers, Final-conv represents a standard convolution block, and FC refers to the fully connected layer.}

\begin{tabular}{|l|l|l|}
\hline
\multicolumn{3}{|c|}{\textbf{Encoder}} \\
\hline
{Layer} & {Input Dimensions} & {Output Dimensions} \\
\hline
Dilated-conv & (B, 1, H, W) & (B, 32, H, W) \\
\hline
Dilated-conv & (B, 32, H, W) & (B, 64, H, W) \\
\hline
ResBlockv1 & (B, 64, H, W) & (B, 128, H, W) \\
\hline
Pool & (B, 128, H, W) & (B, 128, 15, 15) \\
\hline
ResBlockv2 & (B, 128, 15, 15) & (B, 128, 15, 15) \\
\hline
ResBlockv2 & (B, 128, 15, 15) & (B, 128, 15, 15) \\
\hline

\multicolumn{3}{|c|}{\textbf{Decoder}} \\
\hline
{Layer} & {Input Dimensions} & {Output Dimensions} \\
\hline
ResBlockv2 & (B, 128, 15, 15) & (B, 128, 15, 15) \\
\hline
ResBlockv2 & (B, 128, 15, 15) & (B, 128, 15, 15) \\
\hline
Upsample    & (B, 128, 15, 15) & (B, 128, H, W) \\
\hline
ResBlockv1 & (B, 128, H, W) & (B, 128, H, W) \\
\hline
Dilated-conv & (B, 128, H, W) & (B, 32, H, W) \\
\hline
Dilated-conv & (B, 32, H, W) & (B, 16, H, W) \\
\hline
Final-conv & (B, 32, H, W) & (B, 1, H, W) \\
\hline

\multicolumn{3}{|c|}{\textbf{Regression Branch}} \\
\hline
{Layer} & {Input Dimensions} & {Output Dimensions} \\
\hline
Pool & (B, 128, 15, 15) & (B, 128, 8, 8) \\
\hline
ResBlockv2 & (B, 128, 8, 8) & (B, 128, 8, 8) \\
\hline
ResBlockv2 & (B, 128, 8, 8) & (B, 128, 8, 8) \\
\hline
ResBlockv1 & (B, 128, 8, 8) & (B, 128, 8, 8) \\
\hline
Final-conv & (B, 128, 8, 8) & (B, 1, 8, 8) \\
\hline
FC & (B, 1, 64) & (B, 64) \\
\hline
\end{tabular}
\label{table:DTEMDNet_Architecture}
\end{table}

\subsubsection{Hyperparameter Settings for DP-TTA}

In this section, we summarize the hyperparameter settings for the DP-TTA. The optimal combination of hyperparameters was determined through a parameter search strategy. The settings are summarized in the table below:

\begin{table}[htbp]
\centering
\caption{Optimal Hyperparameters for DP-TTA}
\begin{tabular}{ll}
  \toprule
  \textbf{Hyperparameter} & \textbf{Value} \\
  \midrule
  \(\beta_1\) & 1.0 \\
  \(\beta_2\) & 1.0 \\
  Batch Size & 128 \\
  Noise Level & 120.0 \\
  Learning Rate & \( 1 \times 10^{-5} \) \\
  \bottomrule
\end{tabular}
\label{table:hyperparameters}
\end{table}
Besides, DP-TTA adopts a one-step optimization strategy, where the model parameters are updated independently for each batch of test data to avoid the instability often observed in continuous per-sample adaptation.

\section{Experiments}
\subsection{Dataset}
\begin{figure*}[!t]
\centering
\includegraphics[width=\textwidth]{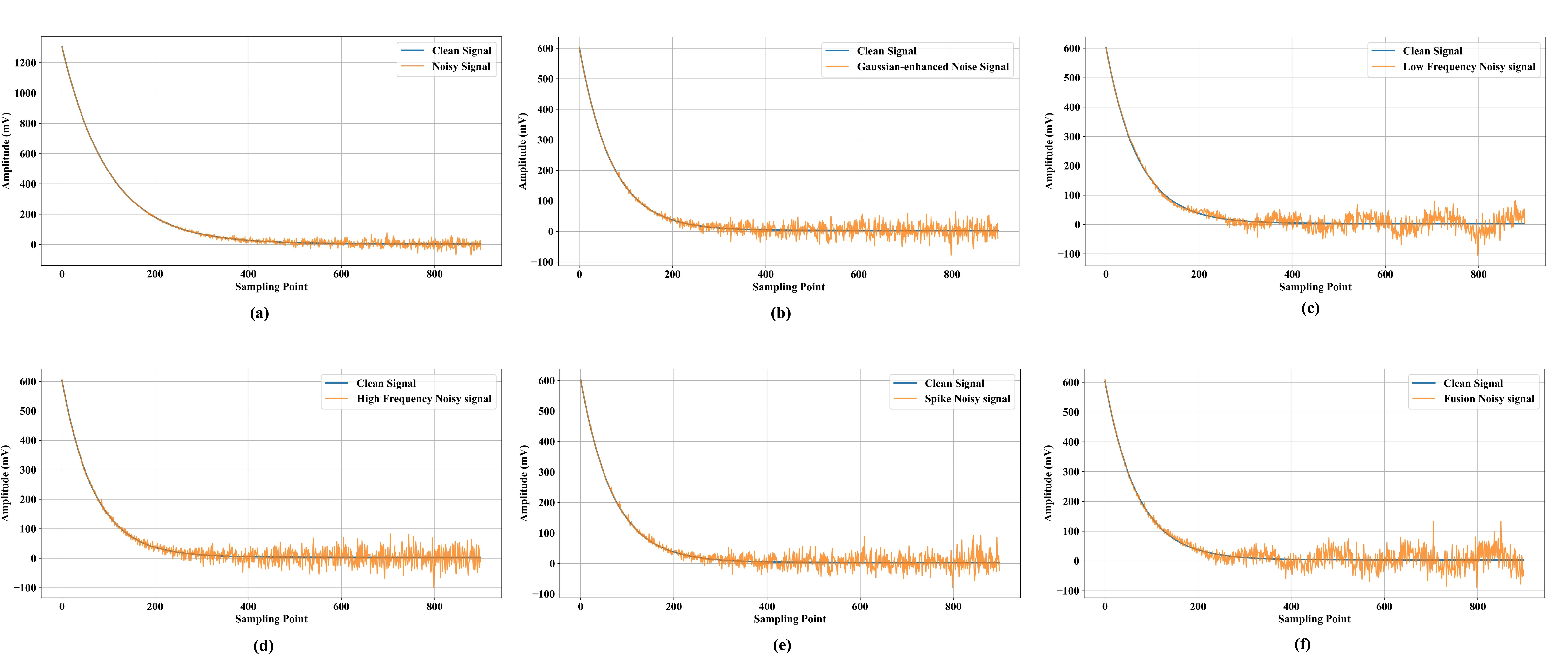}
\caption{Simulation samples. (a) Source dataset sample: \( Q_1 = 1300\), \(Q_2 = 2.5\), \(B = 4.0\), \( SNR = 22\). (b) AGN dataset sample: \( Q_1 = 600\), \(Q_2 =3.6\), \(B = 3.5\),  \( SNR = 10\). (c) LFI dataset sample: \(f = 1.5\), \(A = 30\). (d) HFI dataset sample: \(f = 50\), \(A = 30\). (e) IMP dataset sample: \(k = 30\), \(A_i = 70\). (f) CMP dataset sample.}
\label{fig_3}
\end{figure*}

We prepared both a simulation dataset and a real-world dataset. Fig. \textcolor[RGB]{36,113,163}{\ref{fig_3}} shows some simulation samples.

\subsubsection{Simulation Dataset}
The simulation dataset consists of both a source domain training set and multiple target domain test sets. The clean signal \( s(t) \) is modeled following the theoretical formulation of TEM signals\textcolor[RGB]{36,113,163}{\cite{9258400}}, expressed as:

\begin{equation}
s(t) = Q_1 \sum_{k=1}^{\infty} \exp \left( -Q_2 t \right) + B,
\end{equation}
where \( Q_1 \), \( Q_2 \), and \( B \) represent the amplitude factor, time constant, and Direct Current (DC) offset, respectively. The constants \( Q_1 \), \( Q_2 \), and \( B \) are randomly sampled from the ranges \( Q_1 \in [100, 1500] \), \( Q_2 \in [0.5, 4.0] \), and \( B \in [2.0, 6.0] \), respectively. These ranges are chosen to ensure a diverse set of signal waveforms that realistically capture variations in amplitude, exponential decay behavior, and baseline offset, reflecting different geological conditions and subsurface environments. Once the clean signals are generated, various types of noise are added to form the corresponding noisy signals. All samples in the simulation dataset are constructed following this procedure.

The source domain dataset contains 110,000 samples, divided into training and test sets in a 10:1 ratio. Each sample pair includes a clean signal and a noisy signal, where the noisy signal is generated by adding Gaussian noise with a signal-to-noise ratio (SNR) ranging from 20 dB to 25 dB.

\begin{figure}[!t]
\centering
\includegraphics[width=0.48\textwidth]{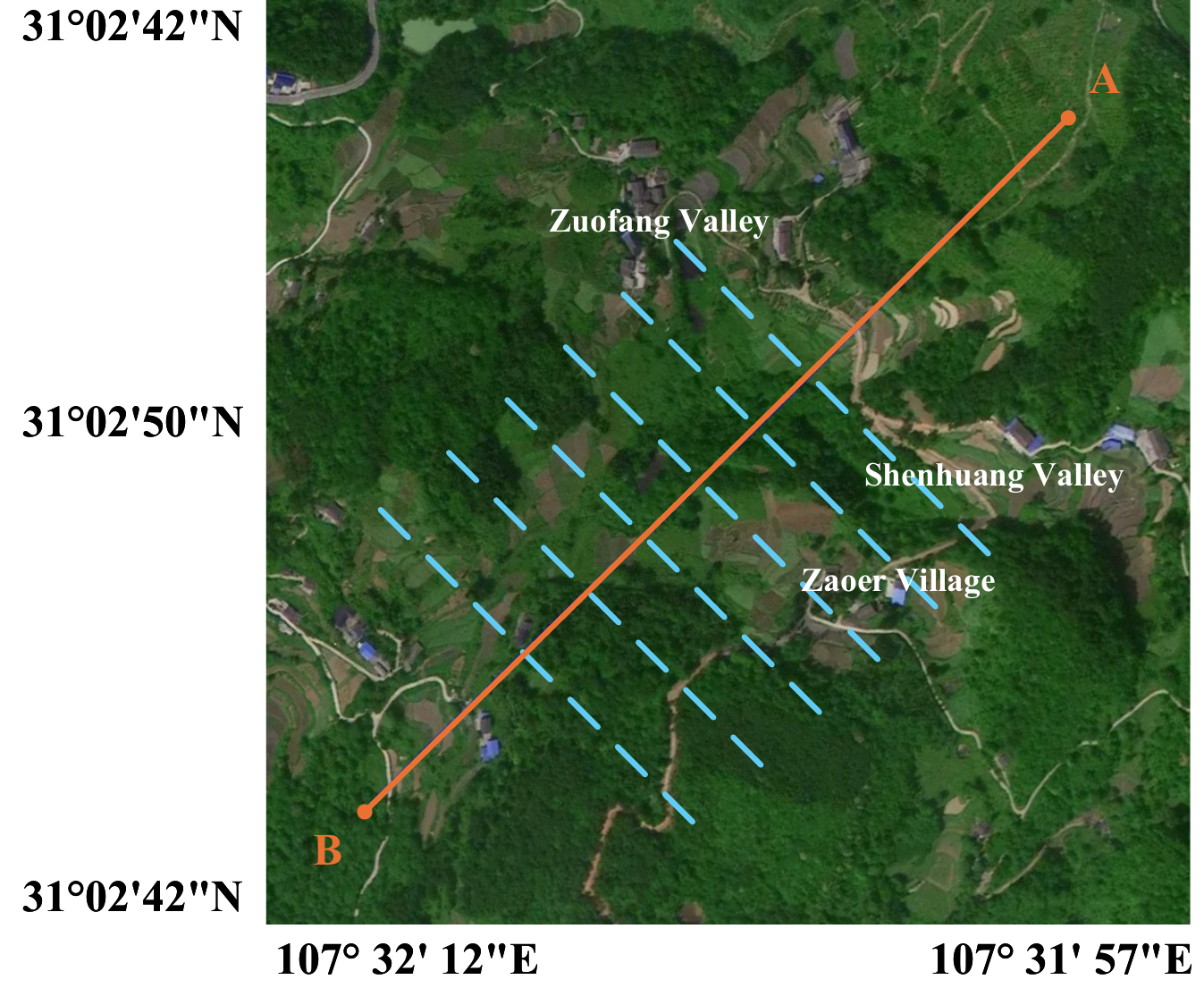}
\caption{Specific location of TEM signal acquisition. Dazhou data acquisition region, with AB length of 0.97 km and the blue dotted line spanning 500 m. The bottom and left panels provide the geographic coordinates of the sampling area.}
\label{fig_4}
\end{figure}

The target domain dataset comprises five subsets, each representing a distinct domain shift scenario and containing 10,000 samples by injecting different types of noise.  Specifically, it includes:

\begin{itemize}
    \item \textbf{Augmented Gaussian noise (AGN) Dataset}: This dataset simulates random environmental noise, typically modeled as Gaussian noise, \(\epsilon(t) \sim \mathcal{N}(0, \sigma^2)\). The noise is added with a higher SNR ranging from 8 dB to 10 dB.

    \item \textbf{Low-Frequency Interference (LFI) Dataset}: This dataset simulates low-frequency environmental vibrations, which are commonly encountered in geophysical surveys due to mechanical disturbances such as drilling activity, sensor shaking, or nearby vehicular movement. The noise is modeled as:
    \begin{equation}
    N_{\text{low}}(t) = A \cdot \sin(2\pi f t + \varphi),
    \end{equation}
    where the amplitude \(A\) is sampled from the range \([10mV, 30mV]\), the frequency \(f\) from \([1 Hz, 5 Hz]\), and the phase \(\varphi\) from \([0, 2\pi]\).
    
    \item \textbf{High-Frequency Interference (HFI) Dataset}: This dataset simulates high-frequency electromagnetic interference, commonly arising from nearby electrical equipment, power lines, or control circuitry in field instrumentation. The noise is modeled as:
    \begin{equation}
    N_{\text{high}}(t) = A \cdot \sin(2\pi f t + \varphi),
    \end{equation}
    where the amplitude \(A\) is sampled from the range \([10mV, 30mV]\), the frequency \(f\) from \([10 Hz, 50 Hz]\), and the phase \(\varphi\) from \([0, 2\pi]\).
    
    \item \textbf{Impulse Noise (IMP) Dataset}: This dataset captures sharp, burst-like interferences frequently observed in field-collected geophysical data due to switching transients, relay operations, or sudden electromagnetic discharges. The noise is modeled as:
    \begin{equation}
    N_{\text{spike}}(t) = \sum_{i=1}^{k} A_i \cdot \delta(t - t_i),
    \end{equation}
    where each amplitude \(A_i\) is sampled from the range \([50mV, 70mV]\), \(t_i\) denotes the temporal location of the spike, and the number of spikes \(k\) is sampled from \([20, 30]\). This type of interference introduces sparse, high-amplitude disturbances that can severely distort transient signal integrity.
    
    \item \textbf{Composite Noise (CMP) Dataset}: This dataset integrates all four types of noise into a single noisy environment, simulating the complex interference conditions typically encountered in real-world field surveys.
\end{itemize}

\subsubsection{Real-World Dataset}
The real-world dataset consists of actual TEM signals collected from a geophysical field survey conducted in the Dazhou region of Sichuan Province, China. Fig. \textcolor[RGB]{36,113,163}{\ref{fig_4}} illustrates the specific location of data acquisition. The orange line AB marks the transmitter positions for electromagnetic emission, while the orthogonal blue dotted line indicates the survey line used for signal recording. The survey lines are spaced at consistent intervals of 100 m, with an effective collection width of 500 m, denoted by the span of the blue dotted line. The data were recorded using our custom TEM receiver with a sampling rate of 250 Hz and a transmission frequency of 25 Hz. Each recording captures 5 seconds of TEM secondary field signals and is subsequently cropped to 900 sampling points per sample.

\subsection{Evaluation Metric}

SNR is a fundamental measure in signal processing that quantifies the ratio of the signal power to the noise power. A higher SNR indicates better signal quality and reduced noise interference. The SNR is calculated using the following formula:

\begin{equation}
    \text{SNR} = 10 \times \log_{10}\left(\frac{\|s_{\text{clean}}\|^2}{\|\hat{s} - s_{\text{clean}}\|^2}\right),
\end{equation}
where \(s_{\text{clean}}\) represents the clean signal, and \(\hat{s}\) is the denoised signal. The numerator corresponds to the power of the clean signal, while the denominator represents the noise power.

\begin{table*}[t]
\centering
\caption{Average denoising SNR (dB) of baseline methods under source and domain shift scenarios. Bold indicates best performance, underlined indicates second best.}
\begin{tabular}{lp{1.2cm}p{1.2cm}p{1.2cm}p{1.2cm}p{1.2cm}p{1.2cm}}
\toprule
\textbf{Model} & \textbf{Source} & \textbf{AGN} & \textbf{HFI} & \textbf{LFI} & \textbf{IMP} & \textbf{CMP} \\
\midrule
Dictionary Learning & 22.36 & 23.71 & 23.83 & 23.74 & 23.64 & 23.70 \\
DnCNN               & 28.61 & 19.29 & 18.77 & 18.41 & 19.03 & 17.76 \\
SFSDSA              & 25.65 & 25.39 & 25.38 & 25.37 & 25.20 & 25.17 \\
ResNet6             & 29.74 & 26.07 & 26.92 & 25.53 & 26.61 & 25.03 \\
TEM1DFormer         & 26.27 & 16.19 & 15.85 & 14.10 & 16.27 & 13.82 \\
TEMDNet             & \underline{36.69} & \underline{27.72} & \underline{27.41} & \underline{26.88} & \underline{27.67} & \underline{26.61} \\
\textbf{DTEMDNet}   & \textbf{40.89} & \textbf{33.52} & \textbf{32.87} & \textbf{32.43} & \textbf{32.82} & \textbf{31.59} \\
\bottomrule
\end{tabular}
\label{table:Baseline SNR results}
\end{table*}

\subsection{Simulation Experiment Results}
\subsubsection{Baseline comparison and analysis}
This comparative experiment is conducted with two objectives: (1) To verify the robustness and adaptability of DTEMDNet in addressing the domain shift problem; (2) To assess the advantage of incorporating dictionary-driven priors over existing TEM signal denoising methods that lack explicit structural constraints.

Accordingly, we evaluate DTEMDNet’s denoising performance by comparing it with several representative baseline TEM signal denoising methods. These baselines include both dictionary learning-based denoising and state-of-the-art deep learning-based methods, namely: Dictionary Learning \textcolor[RGB]{36,113,163}{\cite{aharon2006k}}, DnCNN \textcolor[RGB]{36,113,163}{\cite{zhang2017beyond}}, ResNet6 \textcolor[RGB]{36,113,163}{\cite{zhang2017beyond}}, SFSDSA \textcolor[RGB]{36,113,163}{\cite{lin2019denoising}}, TEM1Dformer \textcolor[RGB]{36,113,163}{\cite{pan2023tem1dformer}}, and TEMDNet \textcolor[RGB]{36,113,163}{\cite{9258400}}. 

For dictionary learning-based denoising, the objective is to minimize the reconstruction error between the dictionary reconstruction and the clean signal. To achieve this, we iteratively optimize both the dictionary and the corresponding sparse codes, enabling effective denoising of the input noisy TEM signals \textcolor[RGB]{36,113,163}{\cite{9528899}}. For TEM1Dformer, we set the input dimension to 1x900, while for all other methods, the input dimension is set to 30x30. Note that the hyperparameter settings of all the aforementioned deep learning denoisers are similar. All baseline DNN denoisers are trained on the source domain dataset for 100 epochs with an initial learning rate of \(1 \times 10^{-3}\) and a batch size of 256. The implementation is based on PyTorch 1.10.1 and runs on a single NVIDIA RTX 4090 GPU. For a fair and consistent evaluation, all methods are tested on the same set of simulation datasets. Table~\textcolor[RGB]{36,113,163}{\ref{table:Baseline SNR results}} presents the denoising performance of the baseline methods across various datasets.

\begin{figure}[h!]
\centering
\includegraphics[width=1.0\linewidth]{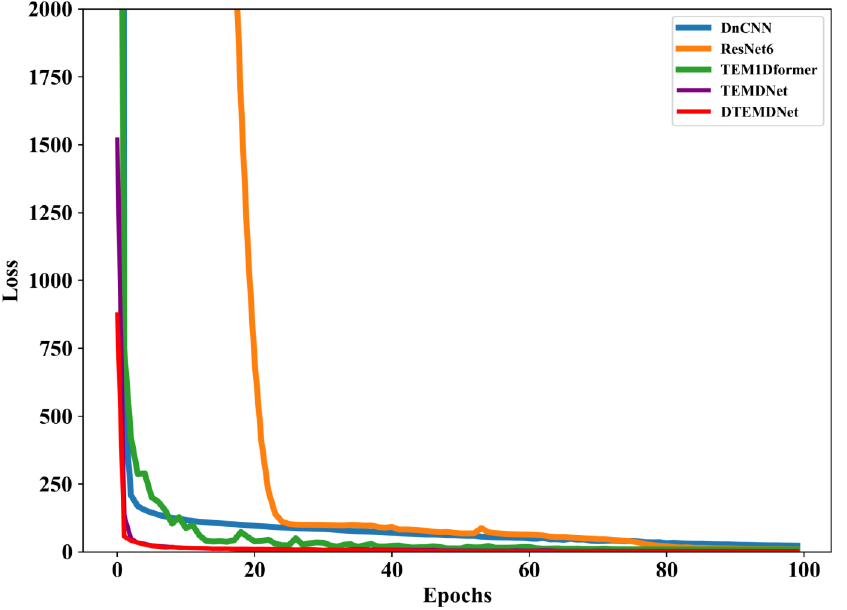}
\caption{Training loss curves of different denoising models on the source domain dataset. DTEMDNet exhibits faster and more stable convergence compared with other baselines, which can be attributed to the guidance of the dictionary-driven prior.}
\label{Model_Loss}
\end{figure}

As shown in Table~\textcolor[RGB]{36,113,163}{\ref{table:Baseline SNR results}}, all DNN-based models experience performance degradation under domain shift. However, DTEMDNet achieves the best overall performance across all test scenarios, with an average gain of 5.19 dB over TEMDNet. While DnCNN and TEM1Dformer perform competitively in the source domain, their generalization degrades significantly under domain shift, with SNR dropping to 17.76 dB and 13.82 dB in the CMP dataset, respectively. SFSDSA and traditional dictionary learning exhibit relatively better stability but fall short in overall denoising fidelity. ResNet6 and TEMDNet maintain solid performance across domains, averaging 26.65 dB and 28.83 dB, respectively. These results provide preliminary validation that incorporating dictionary-driven priors effectively helps the model to tackle the domain shift problem.

In Fig.~\textcolor[RGB]{36,113,163}{\ref{Model_Loss}}, we visually observe the convergence behavior of each DNN-based denoiser. DTEMDNet exhibits the fastest convergence, with the loss rapidly decreasing and stabilizing early, around epoch 10. This is likely due to the incorporation of dictionary-driven priors, which enable the model to adapt quickly to the training data. DnCNN and ResNet6 show slower convergence rates, with their losses remaining high throughout most of the training process. TEM1Dformer and TEMDNet exhibit gradual loss reductions, indicating more stable but slower adaptation. These methods do not achieve the same rapid convergence observed in DTEMDNet, which we speculate is due to the lack of a dictionary-driven prior to guide the learning process.

\subsubsection{Analysis of TTA performance}
In this experiment, several existing test-time adaptation strategies, including Significant-Subspace Alignment (SSA-TTA)\textcolor[RGB]{36,113,163}{\cite{adachi2024test}}, Student-Teacher Test-Time Adaptation (ST-TTA)\textcolor[RGB]{36,113,163}{\cite{vs2023towards}}, and Self-Supervised Test-Time Adaptation (SS-TTA)\textcolor[RGB]{36,113,163}{\cite{fahim2023ss}}, are comparatively studied alongside proposed DP-TTA.
\begin{table*}[t]
\centering
\caption{Average SNR (dB) Results for Different TTA Methods. Note: "Source" represents the pre-trained DTEMDNet model without any test-time adaptation. The values in parentheses for each TTA method indicate the improvement in SNR relative to the Source domain model. Bold indicates best performance, underlined indicates second best.}
\begin{tabular}{lp{1.8cm}p{1.8cm}p{1.8cm}p{1.8cm}p{1.8cm}}
\toprule
\textbf{Method} & \textbf{AGN } & \textbf{HFI} & \textbf{LFI} & \textbf{IMP} & \textbf{CMP} \\
\midrule
\textbf{Source}   & 33.52          & 32.87          & 32.43          & 32.82          & 31.59 \\
\textbf{SSA-TTA}  & 33.87 (+0.35)  & 33.28 (+0.41)  & 32.83 (+0.40)  & 33.16 (+0.34)  & 31.97 (+0.38) \\
\textbf{SS-TTA}   & 34.50 (+0.98)  & 33.89 (+1.02)  & 33.31 (+0.88)  & \underline{34.06 (+1.24)}  & \underline{32.67 (+1.08)} \\
\textbf{ST-TTA}   & \underline{34.68 (+1.16)}  & \underline{34.05 (+1.18)}  & \underline{33.49 (+1.06)}  & 33.84 (+1.02)  & 32.59 (+1.00) \\
\textbf{DP-TTA}   & \textbf{35.47 (+1.95)}  & \textbf{35.00 (+2.13)}  & \textbf{34.32 (+1.89)}  & \textbf{34.77 (+1.95)}  & \textbf{33.57 (+1.98)} \\
\bottomrule
\end{tabular}
\label{table: TTA SNR results}
\end{table*}
SSA-TTA improves the pre-trained model's performance by focusing the alignment on the most important subspace dimensions, rather than attempting to align the entire feature space. ST-TTA is aimed at addressing domain shift in object detection task, which addresses this by leveraging a student-teacher architecture, where the teacher model guides the student model to improve its performance on target domain. SS-TTA introduces a test-time adaptation strategy for both supervised and self-supervised image denoising methods, in which an identity mapping regularization loss is proposed. We use the pre-trained DTEMDNet from the source domain to conduct experiments on target domain datasets. For each TTA strategy, hyperparameters are carefully tuned to achieve optimal performance. To ensure fairness, all TTA strategies adopt a consistent one-step update mechanism, with the average SNR recorded for each dataset.

\begin{figure*}[h!]
\centering
\includegraphics[width=1.0\linewidth]{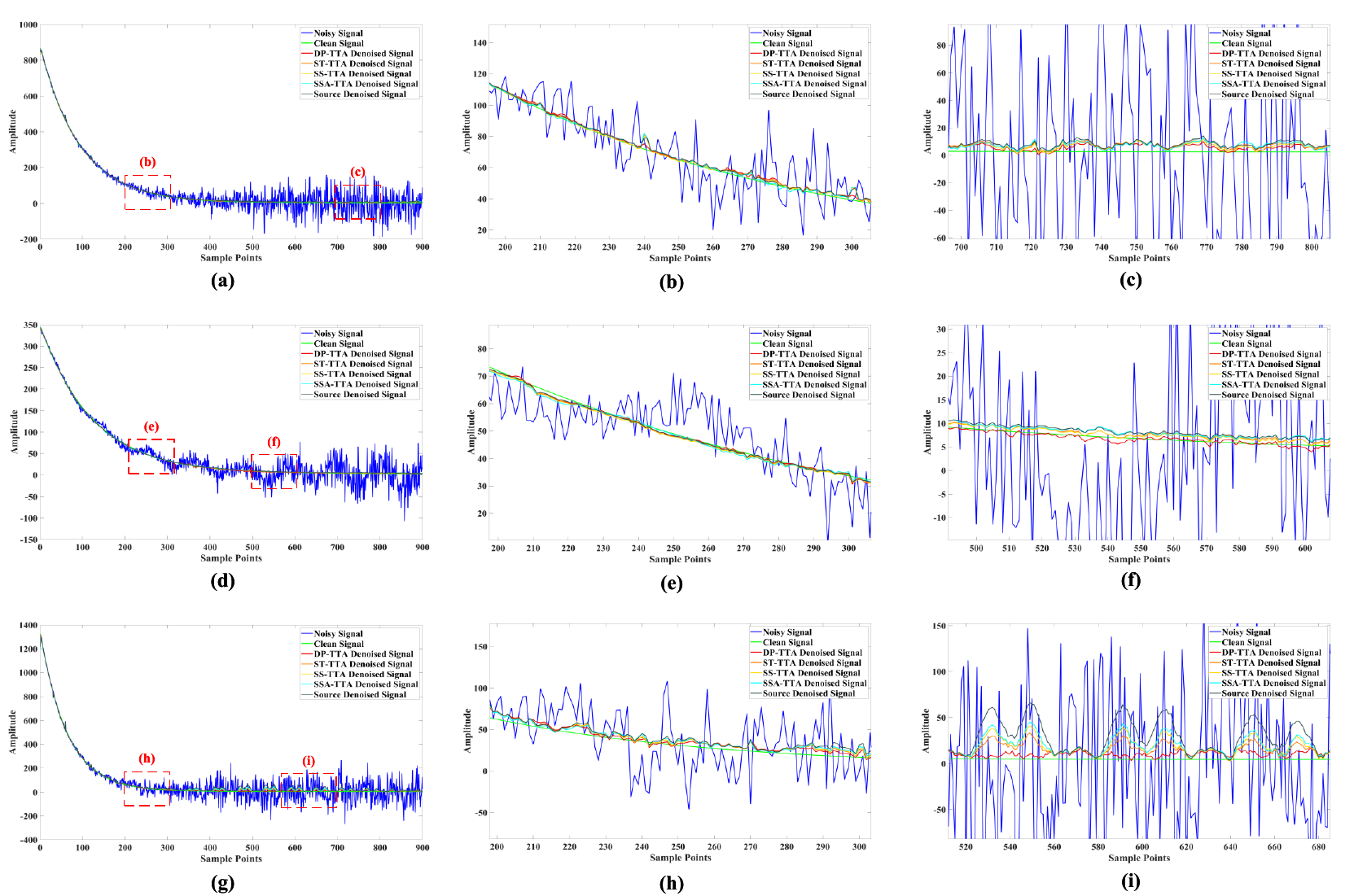}
\caption{Representative denoising results using different TTA strategy. (a), (b), (c) 
represent denoised results from AGN dataset: (a) Overview, (b) Early region zoomed-in, (c) Later region zoomed-in. (d), (e), (f) represent denoised results from LFI dataset: (d) Overview, (e) Early region zoomed-in, (f) Later region zoomed-in. (g), (h), (i) represent denoised results from CMP dataset: (g) Overview, (h) Early region zoomed-in, (i) Later region zoomed-in.}
\label{TTA_denoising_result}
\end{figure*}

Table~\textcolor[RGB]{36,113,163}{\ref{table: TTA SNR results}} presents the quantitative results. All TTA strategies enhance the model's denoising performance, with DP-TTA achieving the highest average SNR gains across all target datasets. Other TTA strategies also show improvements, but they fall short of DP-TTA’s performance. These results suggest that prior-driven methods, such as DP-TTA, may be more suitable for TEM signal denoising tasks.

Fig.~\textcolor[RGB]{36,113,163}{\ref{TTA_denoising_result}} presents the qualitative denoising results of different TTA strategies across the AGN, LFI, CMP datasets. In the early stages of the signal, all TTA strategies show similar performance, with minimal differences in noise reduction. However, in the later stages, a more noticeable divergence is observed. DP-TTA consistently outperforms the other strategies, achieving a clearer reduction in noise and resulting in a smoother denoised signal. As the signal transitions into the low-noise region, distinguishing between noise and signal becomes increasingly nuanced, complicating the denoising process. However, our method continues to perform much better, likely due to the dictionary-driven priors, which enhances the ability to differentiate subtle noise from critical signal characteristics—an area where other TTA strategies struggle.

\subsection{Actual Geological Region Experiment Results}
\subsubsection{Quantitative Results}
Table~\textcolor[RGB]{36,113,163}{\ref{table:Realworld_denoised_result}} presents the denoising SNR (dB) results for different methods across various order numbers. As observed, DP-TTA achieves the best denoising performance, outperforming both baseline models and other TTA strategies. This underscores the strong practical potential of our method in real-world applications. Furthermore, it is evident that all baseline models trained in the source domain exhibit performance degradation due to domain shift, a common challenge faced by many existing TEM DNN-based denoising methods. This emphasizes the need for incorporating TTA strategies into TEM signal denoising tasks.

\setlength{\tabcolsep}{7pt}
\begin{table*}[t]
\centering
\caption{Denoising SNR (dB) Comparison of Different Methods on Real TEM Data. The best performance is highlighted in bold.}
\begin{tabular}{|l|c|c|c|c|c|c|c|c|c|c|c|}
\hline
\textbf{The Order Number} & 1 & 11 & 21 & 31 & 41 & 51 & 61 & 71 & 81 & 91 \\
\hline
\textbf{ResNet6} & 23.82 & 28.39 & 23.32 & 21.48 & 22.41 & 22.91 & 21.92 & 22.70 & 22.31 & 23.69 \\
\hline
\textbf{TEM1DFormer} & 16.27 & 17.38 & 15.04 & 14.63 & 15.11 & 15.41 & 15.09 & 15.31 & 16.35 & 15.58 \\
\hline
\textbf{TEMDNet} & 23.23 & 25.84 & 23.41 & 21.82 & 23.00 & 24.11 & 22.10 & 22.00 & 23.51 & 24.11 \\
\hline
\textbf{DTEMDNet} & 24.81 & 27.80 & 24.08 & 23.78 & 24.01 & 24.32 & 24.38 & 23.97 & 23.76 & 24.91 \\
\hline
\textbf{SSA-TTA} & 24.93 & 27.90 & 24.39 & 23.91 & 24.15 & 24.46 & 24.48 & 24.14 & 24.03 & 25.26 \\
\hline
\textbf{SS-TTA} & 25.71 & 28.86 & 25.28 & 24.83 & 25.04 & 25.54 & 25.56 & 25.14 & 24.97 & 26.11 \\
\hline
\textbf{ST-TTA} & 25.19 & 28.31 & 24.61 & 24.23 & 24.46 & 24.84 & 24.88 & 24.48 & 24.30 & 25.56 \\
\hline
\textbf{DP-TTA} & \textbf{26.20} & \textbf{29.29} & \textbf{25.77} & \textbf{25.26} & \textbf{25.68} & \textbf{26.70} & \textbf{26.14} & \textbf{26.43} & \textbf{26.43} & \textbf{26.93} \\
\hline
\end{tabular}
\label{table:Realworld_denoised_result}
\end{table*}

\subsubsection{Qualitative Results}

\begin{figure*}[h!]
\centering
\includegraphics[width=0.8\linewidth]{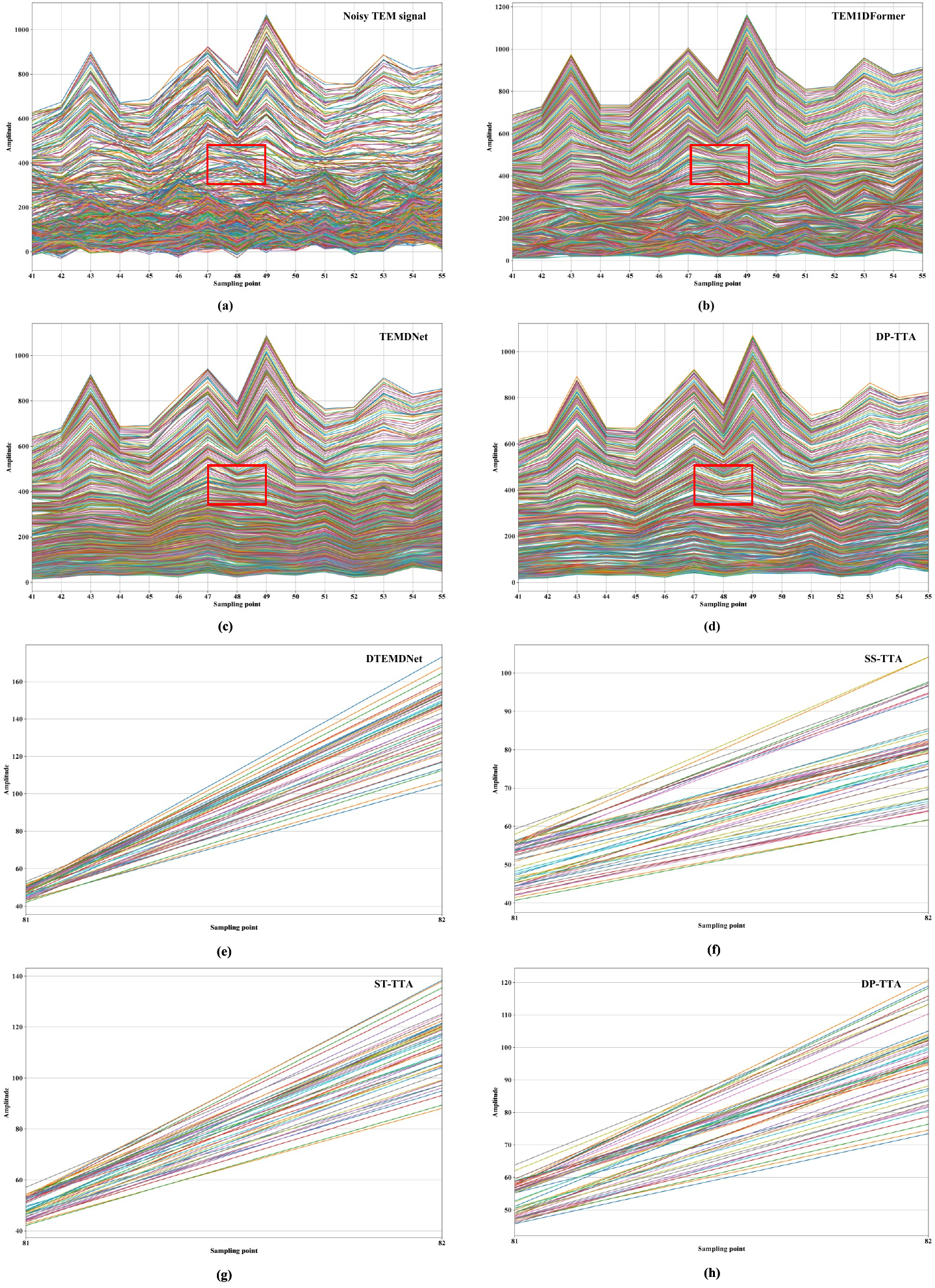}
\caption{Time-domain ordered waveform extraction results on real TEM signals. (a)–(d) illustrate the waveforms from Periods 41st–51nd, comparing baseline denoisers with the DP-TTA strategy. (e)–(h) show magnified late-time waveforms from Periods 81st–82nd, comparing the performance of TTA strategies. The red box is a selected view region for easy comparison. Zoom in for better comparison. (a) Actual noisy TEM signal. (b) Denoising by TEM1DFormer. (c) Denoising by TEMDNet. (d) Denoising by DP-TTA framework. (e) Pre-trained DTEMDNet. (f) SS-TTA. (g) ST-TTA. (h) DP-TTA.}
\label{Realworld_time_wav_image results}
\end{figure*}

The time-domain order waveform image serves as a practical and intuitive tool for qualitatively assessing TEM signal denoising performance. In noisy scenarios, waveform plots often exhibit severe crossovers and high-frequency fluctuations, which obscure key subsurface characteristics and hinder interpretation. In contrast, high-quality denoised signals demonstrate smooth decay, sparse curve crossings, and consistent waveform patterns across neighboring periods—features that are essential for reliable geological analysis. Moreover, late-time signals typically contain richer information about deeper underground structures, making their preservation and clarity particularly important.

Denoised signals from different periods are randomly selected to generate time-domain order waveform images. A qualitative comparison is first performed between existing denoising methods and our method. Subsequently, to better evaluate the performance of TTA strategies, only the magnified view of the later stages of the signals is presented.
Fig.~\textcolor[RGB]{36,113,163}{\ref{Realworld_time_wav_image results}} shows the corresponding results. In Fig.~\textcolor[RGB]{36,113,163}{\ref{Realworld_time_wav_image results}(a)-(d)}, by focusing on the red-box region and the bottom portions of the plots, we observe that DP-TTA yields the cleanest waveform image, with significantly fewer curve intersections and better trend consistency, indicating superior capability in preserving key signal features while effectively suppressing noise.

To further evaluate the effect of TTA strategies, we visualize the denoised waveform images of real TEM signals from periods 81th and 82th, as shown in Fig.~\textcolor[RGB]{36,113,163}{\ref{Realworld_time_wav_image results}(e)-(h)}. These subfigures show zoomed-in views of the late-time signal segments, where differences between methods become more evident. As shown in Fig.~\textcolor[RGB]{36,113,163}{\ref{Realworld_time_wav_image results}(e)}, the time-domain waveform of DTEMDNet exhibits curve convergence and notable crossovers. In contrast, the time-domain waveforms of other TTA strategies appear more uniform and sparse. Although slight crossovers and clustering still exist in all waveforms, our method overall achieves a sparser and more uniform result, highlighting its superior performance improvement.

\section{Ablation study}
In this ablation study, we investigate two key aspects of our method and further explain the motivation behind the introduction of the one-order variation self-supervised loss. Specifically, we focus on: (1) the effect of dictionary atom size, analyzing how different values of $K$ influence the reconstruction accuracy and sparsity of the learned representation; (2) the motivation for one-order variation regularization, examining the structural similarities between the dictionary-reconstructed signal and the clean signal, which underpin the design of the one-order variation loss term; (3) the contribution of each TTA loss component, evaluating the individual and combined effects of three self-supervised losses on final performance.

\subsection{Effect of dictionary atom size}

\begin{figure}[h!]
\centering
\includegraphics[width=1.0\linewidth]{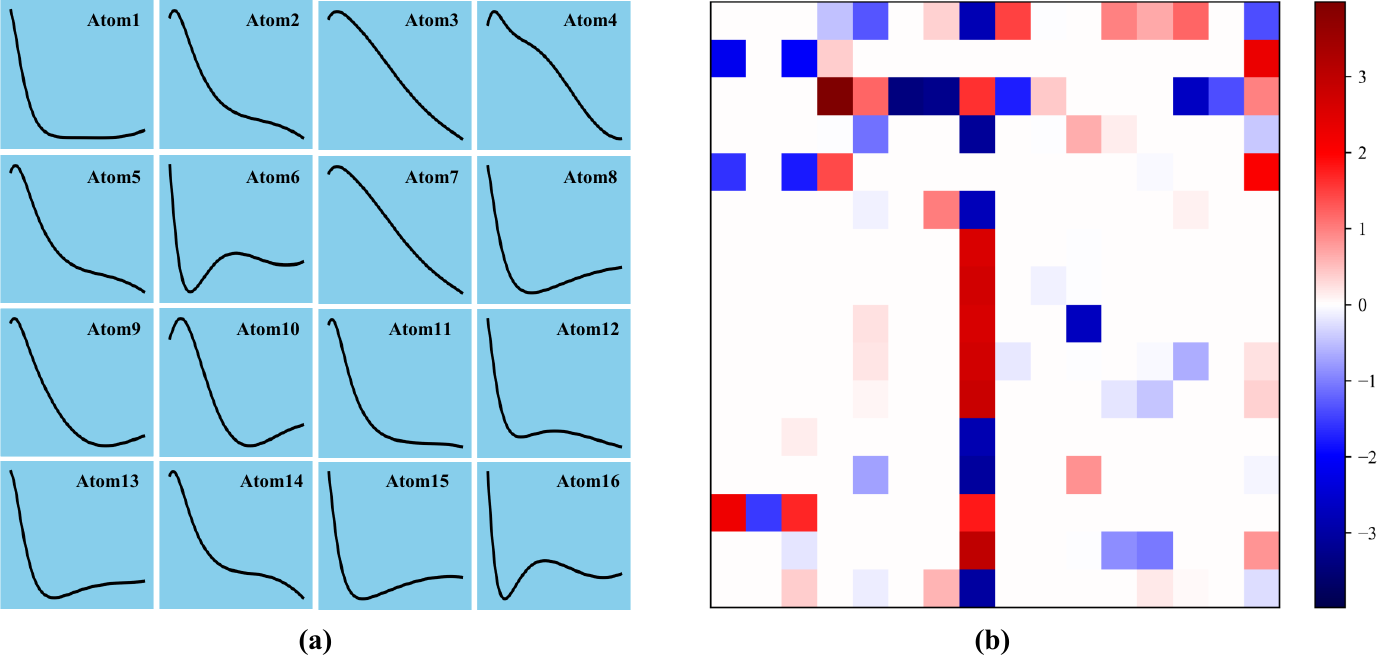}
\caption{A subset of dictionary atoms and the corresponding sparse code when $K=64$. (a) Dictionary atoms, which capture the characteristic temporal decay patterns of TEM signals. (b) Local sparse code, showing how individual atoms are selectively activated to represent different signal components.}
\label{Dictionary atoms and sparse code}
\end{figure}

\begin{figure}[!h]
    \centering
    \includegraphics[width=1.0\linewidth]{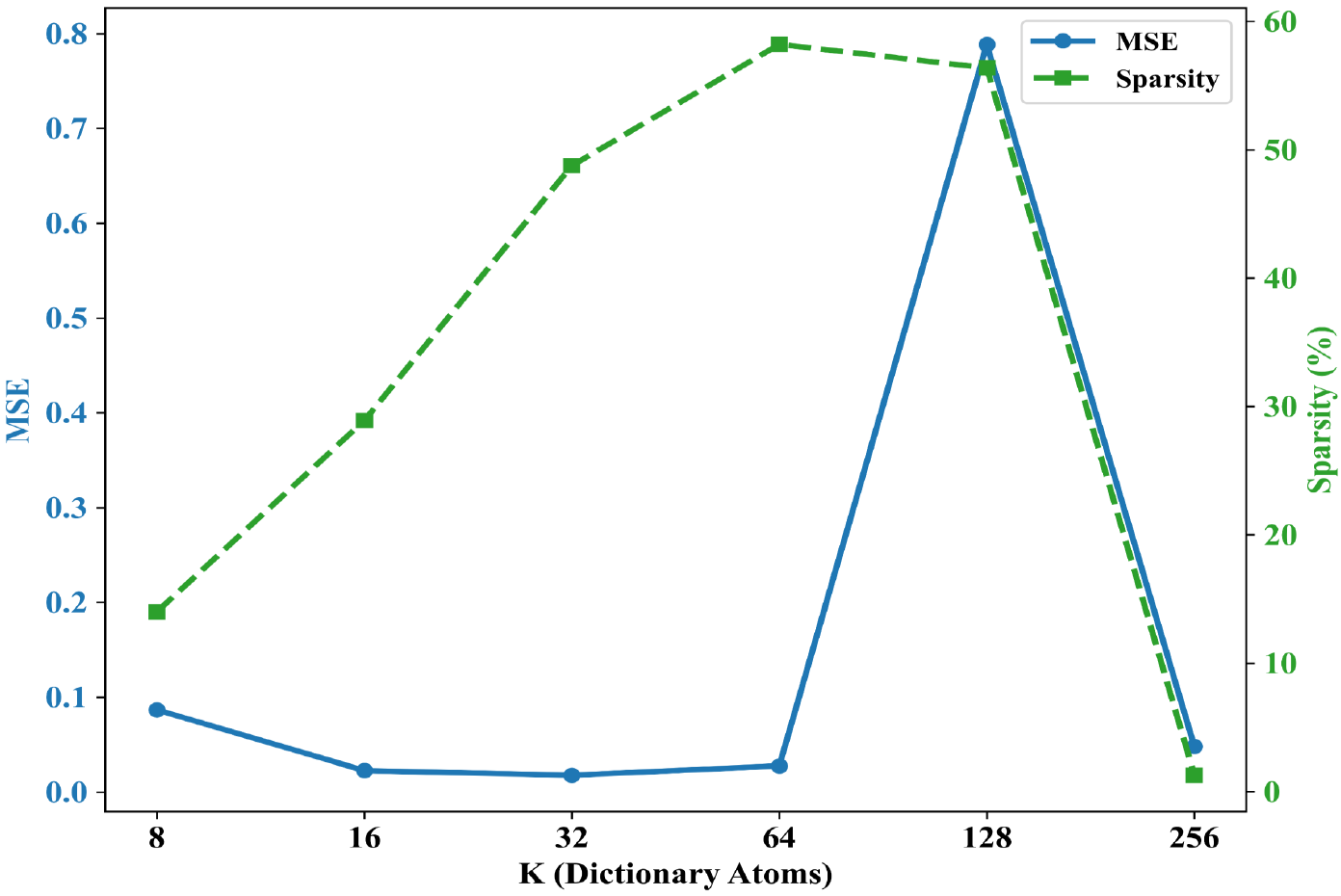}
    \caption{Reconstruction error (MSE) and sparsity for different dictionary atom sizes ($K$). MSE is calculated between the original signals and those reconstructed by the learned dictionary, while sparsity denotes the proportion of zero entries in the sparse coefficient matrix. When $K=64$, the dictionary achieves a good balance between low reconstruction error and high sparsity.}
    \label{DL_Mse_Sparsity_results}
\end{figure}

The quality of the reconstructed TEM signal by dictionary and the sparsity of the sparse are crucial factors determining the effectiveness of the learned dictionary prior. Poor dictionary quality or low sparsity can pose significant challenges for the denoising model and the regression branch, particularly when predicting complex sparse representations.

Fig.~\textcolor[RGB]{36,113,163}{\ref{Dictionary atoms and sparse code}} illustrates a part of dictionary atoms and sparse codes. Intuitively, the displayed dictionary atoms clearly capture the signal's decay and smoothness, while the sparse code consists mostly of zero values, reflecting its inherent sparsity. Fig.~\textcolor[RGB]{36,113,163}{\ref{DL_Mse_Sparsity_results}} reports the reconstruction error (MSE) and the sparsity of the sparse code for different values of $K$. As shown, the MSE decreases as $K$ increases, reaching its minimum at $K=32$. However, further increasing $K$ results in higher MSE, possibly due to overfitting or redundant dictionary atoms. Likewise, the sparsity of the sparse code peaks at $K=64$, but degrades significantly at extreme values such as $K=8$ or $K=256$. These observations highlight the importance of selecting an appropriate $K$ to balance reconstruction accuracy and sparsity, thereby enabling the construction of high-quality dictionary-driven priors for model learning.

\subsection{Motivation for one-order variation regularization}

\begin{figure}[!t]
\centering
\includegraphics[width=1.0\linewidth]{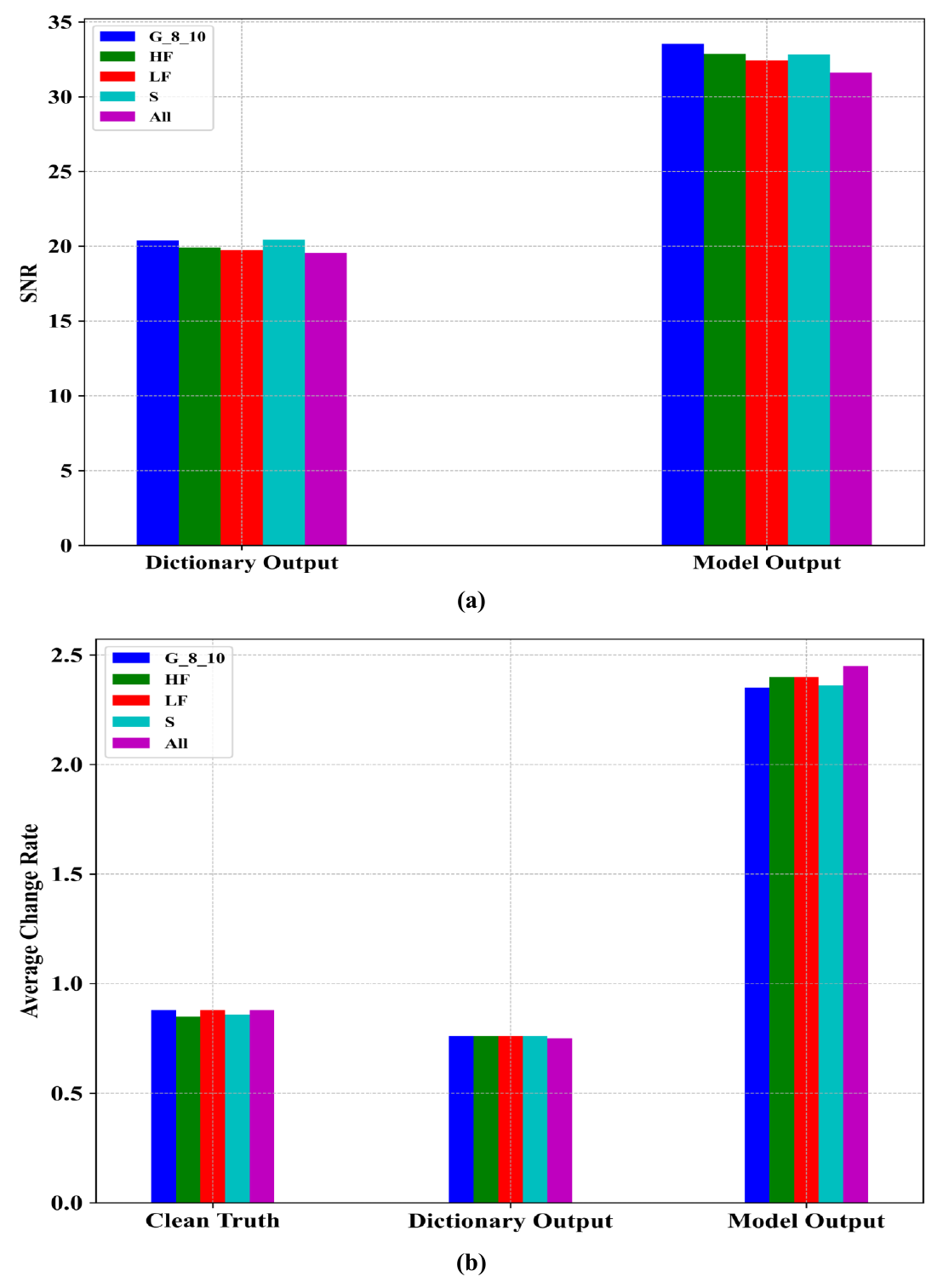}
\caption{Comparison of Model Output and Dictionary Reconstruction on Cross-domain Dataset. (a) SNR Results. (b) Average Rate of Change Results. Note: The average rate of change refers to the mean of the first-order difference for each output result.}
\label{DL_vs_M_Results}
\end{figure}

\begin{figure}[!t]
\centering
\includegraphics[width=1.0\linewidth]{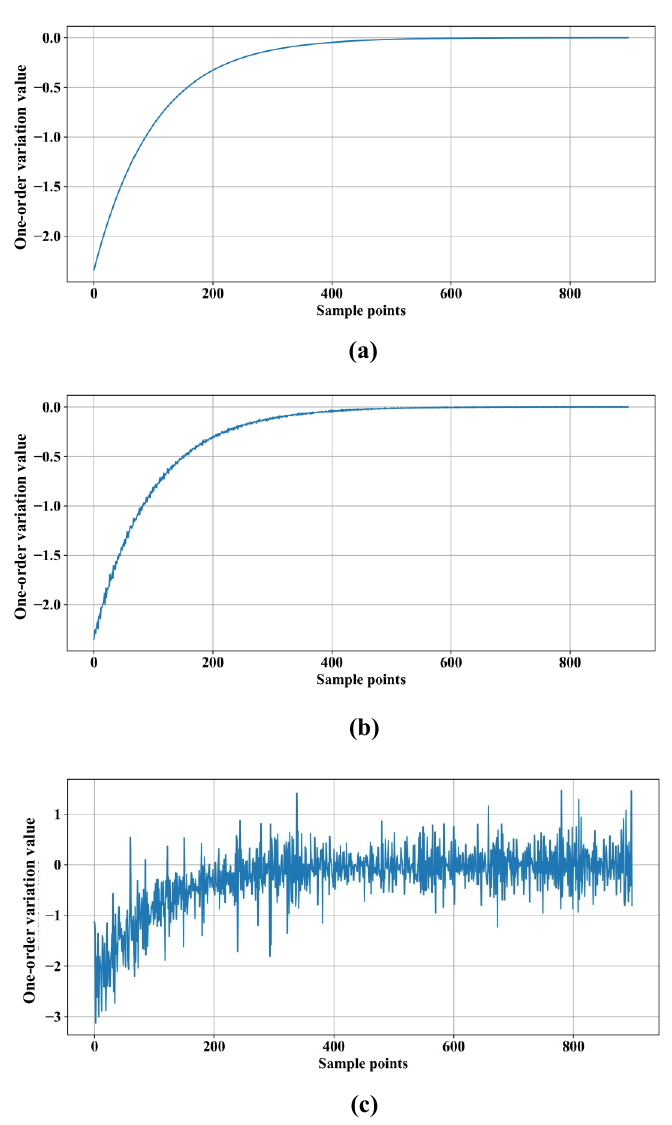}
\caption{(a) One-order variation of the clean TEM signal. (b) One-order variation of the dictionary reconstruction result. (c) One-order variation of the model output result. Note: A random sample from the Cross-domain dataset is input into the model, and one-order variation analysis is performed on the clean signal, dictionary reconstruction result, and the final denoised result. This comparison highlights the difference in one-order variation behavior between the three signals, showing that while the decoder's denoised output yields better SNR, it lacks the smooth one-order variation characteristics of the dictionary reconstruction.}
\label{fig_2}
\end{figure}

We investigated the denoising results of the CNN branch and the dictionary learning branch, and discovered a noteworthy finding. It is readily observed in Fig.~\textcolor[RGB]{36,113,163}{ \ref{DL_vs_M_Results}(a)} that the denoising performance of dictionary reconstruction is consistently inferior to the model's denoising output. Nevertheless, in Fig.~\textcolor[RGB]{36,113,163}{\ref{DL_vs_M_Results}(b)}, the average rate of change (first-order difference) of the dictionary reconstruction signal outperforms the model's output, and is very close to the clean TEM signal, remaining stable throughout. Based on these two findings, we conclude that the dictionary reconstruction results may have a larger overall offset, which leads to higher noise levels. However, it excellently retains the physical characteristics of the TEM signal (exponential decay and smoothness). Fig.~\textcolor[RGB]{36,113,163} {\ref{fig_2}} provides qualitative results. In light of this, a regularization method that provides high-quality labels with physical constraints during the TTA phase naturally arises, namely one-order variation regularization $\mathcal{L}_{\text{one-order variation}}$. This is both an important discovery in our work and the motivation for our one-order variation regularization approach.

\subsection{The Contribution of TTA Loss Component}

Table~\textcolor[RGB]{36,113,163} {\ref{TTA_componet_contribution}} shows the performance improvement with different TTA loss components. Each loss term—\(\mathcal{L}_{\text{denoising}}\), \(\mathcal{L}_{\text{sparse}}\), and \(\mathcal{L}_{\text{one-order variation}}\)—contributes uniquely to the denoising process. The denoising loss focuses on restoring signal fidelity, while the sparse coding term enhances the model’s ability to capture essential features by promoting sparsity. The one-order variation regularization, on the other hand, ensures smooth transitions in the signal, preserving its physical properties, such as exponential decay. Combining these terms results in even better performance, with each component playing an indispensable role in improving robustness and generalization.

\setlength{\tabcolsep}{10pt}
\begin{table}[!t]
\centering
\caption{Performance Changes with Different TTA Loss Components. SNR Increase represents the performance improvement of DTEMDNet on the All dataset with TTA. The best performance is highlighted in bold.}
\label{TTA_componet_contribution}
\begin{tabular}{l c}
\toprule
TTA Loss Component & SNR Increase \\ 
\midrule
\(\mathcal{L}_{\text{denoising}}\) & +0.58 \\ 
\(\mathcal{L}_{\text{sparse}}\) & +0.89 \\ 
\(\mathcal{L}_{\text{one-order}}\) & +1.02 \\ 
\(\mathcal{L}_{\text{denoising}} + \mathcal{L}_{\text{sparse}}\) & +1.65 \\ 
\(\mathcal{L}_{\text{denoising}} + \mathcal{L}_{\text{one-order}}\) & +1.69 \\ 
\(\mathcal{L}_{\text{sparse}} + \mathcal{L}_{\text{one-order}}\) & +1.86 \\ 
\(\mathcal{L}_{\text{TTA}}\) & \textbf{+1.98} \\ 

\bottomrule
\end{tabular}
\end{table}

\section{Discussion}
 The domain shift problem in TEM signal denoising, although largely overlooked in previous studies, is unavoidable in real-world applications. From a bayesian perspective, domain shift leads to model performance degradation due to mismatches in both the likelihood and prior assumptions. To tackle this challenge, we proposed DP-TTA framework, leveraging dictionary learning to extract intrinsic features as the dictionary-driven priors for denoising tasks and guiding TTA strategy. Experimental results demonstrate that DP-TTA achieves superior performance over both baseline and existing TTA methods. Moreover, our model maintains a simple architecture that integrates dictionary-driven priors with CNNs through a regression branch.

Compared to existing TTA strategies, DP-TTA introduces explicit domain-invariant information into the adaptation process, enabling more structured and guided parameter updates. Additionally, the discrepancy between the dictionary-reconstructed signal and the model-denoised output motivates the use of one-order variation regularization, which enforces stronger physical constraints and results in smoother, more consistent signal recovery.

Despite the promising results obtained in both simulation and real-world datasets, certain limitations remain. While DP-TTA demonstrates strong performance in the context of TEM signal denoising, its effectiveness may diminish when applied to other signal types lacking similar physical structures. Moreover, the dictionary construction process introduces additional computational overhead during the initial phase. 

In the future, we plan to explore simpler and more efficient methods to provide prior information to the model, without relying on dictionaries. These methods will then be combined with TTA strategies to achieve better denoising results, while also offering improved generalizability.

\section{Conclusion}

This paper addresses the domain shift issue in TEM signal denoising, explaining its underlying causes from a Bayesian perspective. The DP-TTA strategy is proposed, which combines dictionary learning with test-time adaptation to handle the domain shift problem encountered in real-world applications. Dictionary learning is utilized to provide a more accurate prior for the signals, while TTA leverages this dictionary-driven prior information to perform prediction consistency regularization and adjust the model parameters. Experimental results demonstrate that DP-TTA outperforms both existing baseline and TTA methods, achieving superior denoising performance. Overall, the proposed approach offers a novel solution to the domain shift problem in TEM signal denoising and lays the groundwork for future advancements in robust and adaptive denoising techniques for various real-world applications.

\section*{Appendix A: List of Acronyms}

\begin{table}[h!]
\centering
\renewcommand{\arraystretch}{1.2} 
\setlength{\tabcolsep}{5pt}       
\begin{tabular}{ll}
\toprule
\textbf{Acronym} & \textbf{Full Term} \\
\midrule
TEM     & Transient Electromagnetic \\
DNN     & Deep Neural Network \\
DP-TTA  & Dictionary-driven Prior Regularization Test-time Adaptation \\
EMI     & Electromagnetic Interference \\
GAN     & Generative Adversarial Network \\
EMD     & Empirical Mode Decomposition \\
VMD     & Variational Mode Decomposition \\
WOA     & Whale Optimization Algorithm \\
SMA     & Slime Mould Algorithm \\
WTD     & Wavelet Threshold Denoising \\
SNR     & Signal-to-Noise Ratio \\
CNN     & Convolutional Neural Network \\
FC      & Fully Connected Layer \\
MSE     & Mean Squared Error \\
AGN     & Augmented Gaussian Noise \\
HFI     & High-Frequency Interference \\
IMP     & Impulse Noise \\
CMP     & Composite Noise \\
ViT     & Vision Transformer \\
\bottomrule
\end{tabular}
\end{table}

\bibliographystyle{IEEEtran}
\bibliography{ref}  

\end{document}